\documentclass[pdflatex, sn-basic]{sn-jnl}

\usepackage{lmodern}
\usepackage{url}

\usepackage{hyperref}
\usepackage{subcaption}
\usepackage{array}
\usepackage{siunitx}
\usepackage{multicol}
\usepackage{times,amsmath}
\usepackage{latexsym,amssymb,lscape,multirow}
\usepackage{bm}
\usepackage{csquotes}
\graphicspath{{figures/}}
\jyear{2023}

\begin{document}

\title[ ]{From Mobile Data to Business Insights: An End-to-End Analytics Framework for Large-Scale Urban Mobility Analysis and Decision Support}

\author[1]{\fnm{Thiago} \sur{Andrade}}

\author[1]{\fnm{Shazia} \sur{Tabassum}  %\orcidlink{0000-0003-0782-7054}
}\email{shazia.tabassum@inesctec.pt}

\author[1]{\fnm{Miguel} \sur{E. P. Silva} %\orcidlink{0000-0003-4893-5625}
}%

%\equalcont{These authors contributed equally to this work.}

\author[2]{\fnm{Ricardo} \sur{Dinis}}

\author[1]{\fnm{Jo\~{a}o} \sur{Gama} %\orcidlink{0000-0002-3210-2066}
}%

\affil[1]{\orgdiv{LIAAD}, \orgname{INESC-TEC}, \orgaddress{\city{Porto}, \country{Portugal}}}

\affil[2]{\orgdiv{Mobile Network Analytics}, \orgname{NOS SGPS}, \orgaddress{\city{Lisboa}, \country{Portugal}}}

\abstract{Real-time location data derived from mobile applications is a powerful tool for addressing various urban challenges, including tourism planning, parking management, bus route optimization, and resource allocation. Besides, it offers invaluable insights for shaping strategic decisions in commercial domains such as location-based services, market share analysis, and behavioral profiling. In this expansive study, we aim to address all of the aforementioned challenges by investigating the behaviors and patterns of smartphone users within urban environments, particularly in the domains of tourism, transportation, and retail.

Our approach encompasses the development of a sophisticated data platform from inception to implementation, which includes the formulation of use cases, architectural design, and implementation of modules. We employ state-of-the-art techniques and technologies, including data anonymization, ETL pipelines, and utilizing Google BigQuery and Vertex AI for data processing and machine learning model development. A modular architecture based on reusable analytical building blocks was developed to generate data products that support multiple stakeholder-driven use cases. Additionally, we apply interactive data visualization techniques via Power BI to facilitate the effective interpretation of analytical findings by stakeholders.

The developed models address a wide range of mobility analytics tasks, including mobility profiling, frequent trajectory mining, area-of-influence analysis, traffic anomaly detection, and origin–destination pattern analysis. The results demonstrate the framework's ability to capture user mobility dynamics at fine spatial and temporal resolutions, providing actionable insights for urban planning and strategic business decision-making. Overall, the proposed work offers a scalable and reusable solution for transforming raw mobility data into operational intelligence, supporting informed decision-making across urban planning, tourism management, transportation systems, and commercial analytics.}

%\newpage
%\tableofcontents
\maketitle

\section{Introduction}

Mobile communications are an essential pillar of a digital society, with smartphones, in particular, as a mandatory accessory for most of the population, as both their main means of communication and point of access to digital content. Society's reliance on smartphones makes these devices an extension of the self, enabling user profiling from the data associated with the phone and its communications. Mobile phone operators are then in a privileged position to exploit this information through technology embedded in communication networks that fulfil services required by users. Their access to mobile communication data, specifically device location, enables mobile phone operators to capture, store and ubiquitously analyze massive amounts of data to generate insights on user patterns and profiles. These insights have a myriad of applications, from social good applications (in case of natural catastrophe, they can help civil authorities in planning evacuation routes or understanding tourist movement patterns to improve tourism routes and identify points of interest) to being sold as data monetization services (for commerce, banking or insurance companies to understand client movement patterns).

The pervasive reliance of society on mobile phone communications means that the quantity of mobile communications data generated per day is massive, imposing constraints on how it is stored, processed and analyzed as traditional methodologies are ill-equipped to handle such massive datasets. Another important aspect of mobile communication data is its streaming quality. New data is constantly being generated, and analytic models often have to answer questions in real time in this data stream. Over the past decade, there has been a large research effort to handle all facets of the ever-increasing volume of data generated worldwide. However, there is a constant need for novel methodologies for extracting knowledge from massive data streams, and existing solutions have often faced challenges to be adapted at scale for industry. It is within this context that the \emph{City Analyser} project was proposed in Portugal, a consortium involving INESC-TEC -- a research institute --, NOS -- a Portuguese telecommunications and media company --, SONAE MC -- a commerce company -- and TUB -- a public transportation company in Braga, Portugal.

The main objective of City Analyser was to develop a data platform to determine behaviours and habits of smartphone users, creating a client profile that enriches population-level analysis of movement patterns in urban environments. This data platform is the supporting environment for analytics focused on four main questions: area of influence of a particular location (where are the clients who visit that location coming from), mobility profiles (identification of city hot spots for social gatherings and mode of transportation when travelling), most frequent trajectories (identifying the path taken to travel between two locations) and anomaly detection (generate alerts when user behaviour shifts from observed normal patterns). The contribution describes the methods developed to answer these four questions.

%Our approach - building blocks and use cases.
%Summary of contributions.
To achieve the outlined objectives and address the specified use cases, we have adopted a modular architecture known as "building blocks." Each building block represents a model designed to tackle a distinct task, such as user residency status identification or mode of transport classification. These building blocks are equipped to dynamically learn, predict, and populate temporary tables. 
They are subsequently queried or directly invoked by use case scripts to generate output tables, referred to as data products. These data products serve as the basis for generating insightful results and graphical representations.

While existing research in the field of mobility pattern mining often focuses on addressing specific or a limited set of related problems, typically narrowed down to specific issues. The research works enumerating various applications and methodologies are predominantly survey papers, often lacking practical implementations \cite{atluri2018spatio}. In contrast, our work is applied research, presenting a comprehensive exploration of diverse problems across three domains. Our contributions are multi-fold, as given below:

\begin{itemize}
    \item The development of a sophisticated data platform with a holistic approach to data analysis, from use case formulation to module implementations, facilitates effective data processing and modeling.
    \item Utilized BigQuery for efficient and rapid data processing, retrieving model outputs and executing use cases.
    \item The models are adaptable to produce outputs tailored to various tessellations, including S2 cells \footnote{http://s2geometry.io/} (grid system for indexing geographies developed by Google), sections (NUTS), and municipalities.
    \item Applied various machine learning techniques tailored to specific tasks, such as rule-based models for user profiles, transfer learning using Random Forest Classifier and Multilayer Perceptron for mode of transport classification, normalized Z-score for temporal anomaly detection of traffic flow, OD matrices, Ramer-Douglas-Peucker algorithm for simplifying trajectories, and DBSCAN with discrete Frechet Distance for clustering frequent trajectories.
    \item Addressed multiple challenges, including stay locations analysis, traffic flow analysis, catchment area analysis, market share of time analysis, and night stay area analysis, leveraging the models above.
    \item Created dashboards to visualise and analyse outputs resulting from data products.
\end{itemize}

The remainder of this paper is organized as follows: Section 3 reviews research works related to the issues addressed in this paper, divided into several topics of relevance given the multi-faceted nature of this study. Section 4 provides an overview of the architecture of the data platform, detailing the data flow, pipeline, and tools utilized, with Section 4.2 offering a comprehensive description of the dataset. Section 5 outlines the pre-processing steps and presents the fundamental definitions necessary for understanding the subsequent sections. All models developed and analyzed in this study are presented in Section 6, with detailed explanations of their structure and function. Section 7 discusses various case studies and demonstrates the application of the building blocks in real-world scenarios. Finally, Section 8 summarizes the findings and conclusions of this research, highlighting the key contributions and potential areas for future work.

\section{Related Work}
\label{relatedwork}
The utilization of mobility data has proven invaluable in tackling numerous real-world challenges. It has facilitated in solving issues like predicting infectious disease outbreaks \citep{kraemer2020effect}, investigating crimes \citep{gerber2014predicting}, protecting wildlife and environment \citep{cagnacci2010animal}, modelling human movements in natural disasters \citep{han2019cities}, and measuring the severity of air pollution \citep{shaji2022study}. Besides these, a vast range of literature focuses on leveraging mobility patterns for shaping the evolution of smart cities and urban development \citep{steenbruggen2015data}. Data-driven strategies have been explored for planning, managing and optimizing transportation systems \citep{zhu2018big, mahrez2021smart} fostering sustainable urban development. The nuanced interplay between human movement and urban dynamics has been extensively leveraged to transform cities into intelligent, responsive ecosystems. This focus aligns with the primary objective of our research. Therefore, we delve into relevant literature within these domains below, highlighting specific models and methodologies to tackle diverse tasks that are further investigated here. The classification is based on the nature of models used in trajectory analysis and predictive analytics.
\subsection{Trajectory Analysis}
\textbf{Cluster mobility and profiling:} The default approach can be simply distance-based algorithms \citep{ben2019clustering}, clustering-based models \citep{atluri2018spatio} that focus on extending traditional clustering algorithms such as K-means, BIRCH, DBSCAN, OPTICS, and STING or collective motion arrest. These models are usually sensitive to variation in spatiotemporal scale and miss the length variant and irregular trajectories \citep{liao2005clustering}. \citep{yue2019detect} proposed DETECT that handles variable-length input by adapting an autoencoder trained using a large volume of unlabeled trajectories. Another class of algorithms deals with uncertain trajectory data where the object moves continuously. At the same time, the location is only recorded at discrete times \citep{yuan2017review}, which happens in the case of RAN Geolocation data.

\textbf{Anomaly detection:} Most anomaly detection works focus on clustering and point densities. More recently, tensor-based methods have been introduced as anomaly detection techniques. We have recently investigated the application of event detection tensor decomposition to dynamic O/D data using a hybrid tensor model called HTM \citep{fanaee2016event}. \cite{sun2017dxnat} detected non-recurring anomalies caused by real-time disruptions or events such as sports, weather, and accidents using Convolutional Neural Network (CNN). Similar strategies can be applied to avoid QoE (Quality of Experience) degradation in the mobile network.

\textbf{Modeling catchment areas:} Some techniques apply gravity models to analyze consumer behaviour and approximate the catchment area of a store or shopping centre by considering the spatial distribution of competition areas and assessing their attractiveness for different population groups \citep{dolega2016estimating} based on some location factors such as the population of the city, the price and supply of the products offered. Some other models include entropy maximization \citep{daniotti2023maximum}, competitive destinations model \citep{cronje2020review}, multi-purpose shopping model for estimating market share \citep{drezner2023multipurpose}, the travel-to-store model \citep{pratt2014delineating}, zone design problem \citep{kohani2012exact}.

\textbf{Characterizing users based on mobility patterns:} Another application relevant to our case study is characterizing users into nationals (intra-region and inter-region), foreigners and tourists based on their mobility patterns. \cite{xu2021towards} introduced nine mobility indicators, such as duration of stay in a city, the spatial extent of activities, the location visited, and trips conducted, and mobility diversity to capture different facets of tourist travel behaviour. Further, they applied an eigen-decomposition approach and Principal Component Analysis to understand the variations and dependencies between the above indicators for individual travellers.

\subsection{Predictive Analytics}
\textbf{Mobility prediction:} This task encompasses problems like next location prediction, next trip, or trajectory prediction \citep{andrade2024we, ma2022individual}. Most works in the area are focused on predicting the mobility of a single user using different approaches such as Markov models \citep{andrade2024we,lv2016big}, non-linear time series analysis \citep{aljeri2020performance} and dynamic Bayes networks \citep{hou2016predicting}. However, the techniques mentioned above often fall short of accurately modelling multidimensional data and predicting the mobility patterns of multiple users concurrently. Advanced methods like tensor models and matrix factorization have been applied in multivariate spatiotemporal analysis and collaborative mobility prediction to address this complexity. For instance, the tensor-based prediction framework introduced by \cite{bahadori2014fast} can incorporate various properties in spatio-temporal data. However, tensor methods suffer from a cold start problem. Additionally, a crucial area of exploration involves real-time mobility prediction. An online CNN model is implemented for trajectory prediction in the work by \cite{ouyang2016deepspace}. In another notable study by \cite{fattore2020automec}, the authors propose a distributed structure for mobility prediction by using the Long Short-Term Memory model (LSTM).

\textbf{Predicting traffic bursts and hotspots:} Road-level traffic prediction is a similar problem as above that can be modelled as a time series of traffic flow data. Lately, RNN and LSTM are popularly used for time series spatiotemporal data prediction \citep{wang2020deep}. It is closely related to the problem of real-time event detection. It can be exploited for decision support in scenarios such as optimizing outdoor parking spaces and resources at supermarkets or opening new stores. \cite{sole2016model} modelled traffic flow as a complex network and used the betweenness measure to analytically predict congestion hot-spots.

\textbf{Transportation mode classification:} Detecting the transportation mode of a trip is not a new problem, quite several works have investigated this problem and are still being benefited by the advancement in learning algorithms and quality of data. \cite{huang2019transport} presented a systematic review of transport mode detection based on mobile phone network data. Most of the traditional methods are rule-based where features like speed, duration and distance are used to categorize transport mode. However, \cite{dabiri2018inferring} says hand-crafted features have drawbacks including vulnerability to traffic and environmental conditions, therefore, they take advantage of CNN architectures which automatically drive high-level features and predict travel modes based on only raw GPS trajectories, where the modes are labeled as walk, bike, bus, driving, and train.

\textbf{Analysis and estimation of Origin/Destination flows:} An OD matrix represents the flow of different entities from a set of origins to a set of destinations. The estimation of O/D matrices has been one of the primary concepts in transportation. Extensive research is carried out in this area to efficiently predict and estimate O/D matrices. A nuanced facet of this challenge involves the estimation of time-dependent O/D matrices. In \cite{djukic2012efficient}, the authors studied the use of Kalman filters to estimate O/D matrices.  \cite{krishnakumari2019data} explores a data-driven approach for this purpose. \cite{moreira2016time} applied an incremental algorithm to discretize the target variable’s historical values on each matrix cell. \cite{ou2019learn} employs a CNN-based model to learn the patterns from dynamic mappings between time-varying O-D flows. However, a significant challenge lies in the sheer size and high dimensionality of these matrices.

\textbf{POI recommendation:} Points of interest recommendation have emerged as one of the popular sub-problems in the area of recommendation with a plethora of works focused on this issue. Most of these works are concentrated on Location-based online social networks \citep{islam2022survey}. However, they can be applied to mobility data gathered from any relevant source. This problem is very close to the above mobility prediction, except it focuses on the user level. The recommended POI or the so-called next place can be a known or unknown place to that user. Nevertheless, these personalized recommendations are essentially based on historical time series data such as the user's past visits, and location preferences and can leverage contextual information such as the user's current location, time of day, type of POIs, neighbourhood or explore collaborative approaches like friendship networks, similarity measures and clustering algorithms to recommend new places based on the trajectories of other users with similar mobility patterns. \cite{wang2020predictability} used the historical trajectories of a user to infer his future location. To that end, they employed a high-order Markov chain model to predict the most likely locations visited by each user. \cite{kong2018hst} applied HST-LSTM (Hierarchical Spatial-Temporal LSTM) to predict an individual’s short-term next location. \cite{liu2021attention} proposed a category-aware gated recurrent unit (CA-GRU) model to decrease the impact of sparse check-in location data from social network services. 

\section{Architecture of data platform}
\begin{figure}[!h]
\centering
\includegraphics[width=1\textwidth]{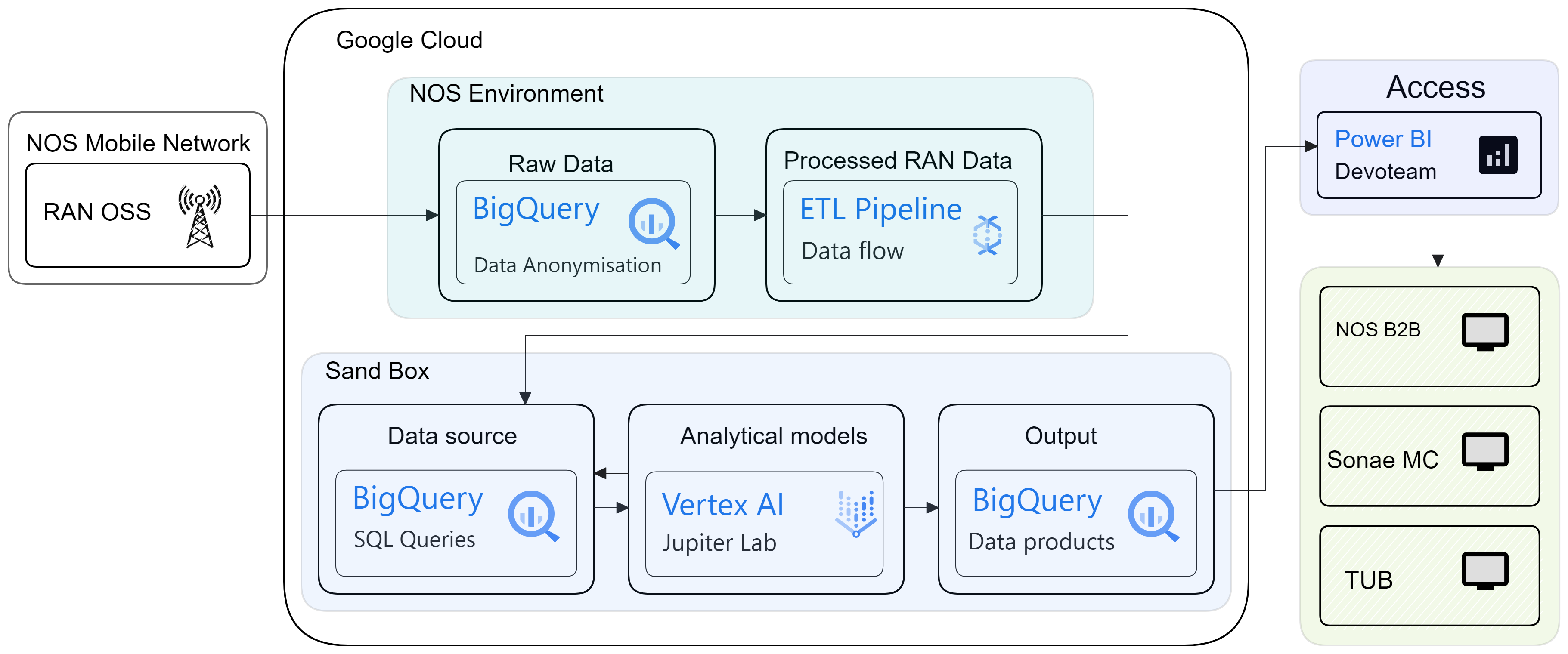}
\caption{Architecture of Sandbox in Google Cloud }
\label{fig:sandbox}
\end{figure}
\subsection{Development Environment}
A working environment was provided by NOS for the research of analytical models by the INESC team. This environment is a sandbox that allows data exploration, model training and producing data products by harnessing the power of Python in Vertex AI, which is provided on GCP (Google Cloud Platform) \citep{gcp} as a complimentary next generation technology. A brief overview of the sandbox is given in Figure \ref{fig:sandbox} below. RAW data from NOS Mobile Network's (another service area) Operation Support Systems (OSS) is fed into NOS project environment. This environment in Google cloud where the RAW data is processed and Anonymised (using techniques below) is fully isolated from the Sandbox provided to develop models. This strict isolation also ensures that there is no possibility of retrieving or downloading data, thereby safeguarding both privacy and security. Jupiter Lab integrated in Vertex AI facilitates querying and fetching data from only permitted Big Query tables using Python files. A range of these Python libraries are used to build machine learning models on the cloud console. Outputs will be accessed by all the project partners with specific permissions. Additionally, ingested by Power BI to generate reports and dashboards. Below we discuss in more detail the capabilities and functions of modules in this architecture.
\subsubsection{Data Anonymisation}
%[5:24 PM] Ricardo Jorge Dinis
Any equipment connected to NOS's mobile network sends and receives data from the network, which depends on the equipment's action, from a telephone call to the sending of an email.
This data contains technical information, such as the antennas to which the equipment is connected. Based on this data, and using mathematical models (such as trilateration) it is possible to calculate, among other things, the longitude/latitude of the equipment.
Because the nature of the data is very sensitive, it is necessary to filter and anonymize the data before it is made available:
Any information that allows users to be identified (number, type of equipment, etc.) is either discarded or processed in such a way as to make the information unreadable;
Users who are in rural areas, without at least X number of people nearby, are discarded, since, although anonymized, the date allows us to discern who the person is by their simple position;
Users who are outside the areas of interest (areas A andB), or outside the period under study (between X and Y), are discarded.
The final result was a source of anonymized atomic data (long/lat) of NOS customers, which are within the areas of interest, over time. In practice, for each instant of time (seconds), we can see the location of the aggregate of customers that are within the areas of interest, but thanks to the anonymization made, it is impossible to see the location of a specific customer.
With this data source, although it is impossible to identify an individual, it is possible to study population flows, main axes used, destination origins, population density, etc.

\subsubsection{ETL Pipeline}
Set up to run every day to use the dataset with the mobility information of mobile network users. Google Dataflow \citep{dataflow} is a fully managed service that modifies and enhances data in both batch (historical) and stream (real-time) modes.
\subsubsection{BigQuery}
BigQuery \citep{bigquery} is a serverless, highly scalable, and cost-effective cloud data warehouse that allows fast queries at petabyte scale. BigQuery’s serverless architecture decouples storage and computing and allows them to scale independently on demand. This structure offers both immense flexibility and cost controls for customers because they don’t need to keep their expensive computing resources up and running all the time. This is very different from traditional node-based cloud data warehouse solutions or on-premise massively parallel processing (MPP) systems. This approach also allows customers of any size to bring their data into the data warehouse and start analyzing their data using Standard SQL without worrying about database operations and system engineering.
\subsubsection{Vertex AI}
Vertex AI is a machine learning platform provided by Google Cloud \citep{vertexAI}. It is designed to help users build, deploy, and manage machine learning models at scale. Vertex AI offers integration with BigQuery, a fully managed, serverless data warehouse provided by Google Cloud (read more above). This integration enables users to leverage the capabilities of both platforms for advanced data analytics and machine learning tasks. Once models are trained, Vertex AI allows users to deploy them to production environments for inference. It supports serving models through REST APIs and provides features for monitoring model performance and health.  Google Cloud's Vertex AI platform offers integration with Jupyter Notebooks, providing with a flexible and interactive environment for developing, experimenting with, and deploying machine learning models. The models are developed in this environment using notebooks and python scripts.  
\subsubsection{Power BI}
The interactive data visualisation software Power BI \citep{powerbi} is used to visualise and interpret use case results and statistics. The access to output from the cloud to Power BI is restricted exclusively to authenticated service accounts, ensuring secure access. Subsequently, the reports and dashboards generated through this process are made available to partners through designated user accounts.

\subsection{Dataset}
\label{subsubsec:PTelco}
%%%%%%%%%%%%%Dataset%%%%%%%%%%
This is a new dataset based on mobile phone data. 
The dataset contains 651.503 instances from 9 months or 250 days starting in March 2022 and finishing in November 2022, consisting of 466 different individuals. 
After cleaning and removing the duplicates it was reduced to 534.612 instances. 
The points were recorded in the Lisbon, Portugal area with a mixed granularity of sampling with more logs recorded as the users were using the network services such as streaming video, social networks, SMS or calls, and also some heartbeat for the telecom systems. 
No information about the users is derived from these data as the entire dataset is anonymized using personally identifiable information (PII) is any information connected to a specific individual that can be used to uncover that individual's identity, such as their social security number, full name, email address or phone number. 
Each point consists of a user sequential identification number, a pair of (latitude, and longitude), and a timestamp. 
A mobile country code was also provided but discarded as we did not use data other than the location points for this work. 
The data was delivered in two files and due to GDPR reasons, we can not make it public.

\subsection{Data Monetization Models}
The analytical models aimed to address the requirements of use cases defined in the project. These product models are indicated as building blocks are outlined in the next section. In the figure below we delineate the data analytics architecture. The data is updated in BigQuery tables from where the Data preprocessing step (Section \ref{datapreprocess} fetches data using Queries and then cleans and preprocesses. The data frames created from the data are used for model training and testing. Use cases call models or query the results stored in temporary Big Query tables which keep on updating on sliding window.

\begin{figure}[!h]
\centering
\includegraphics[width=1\textwidth]{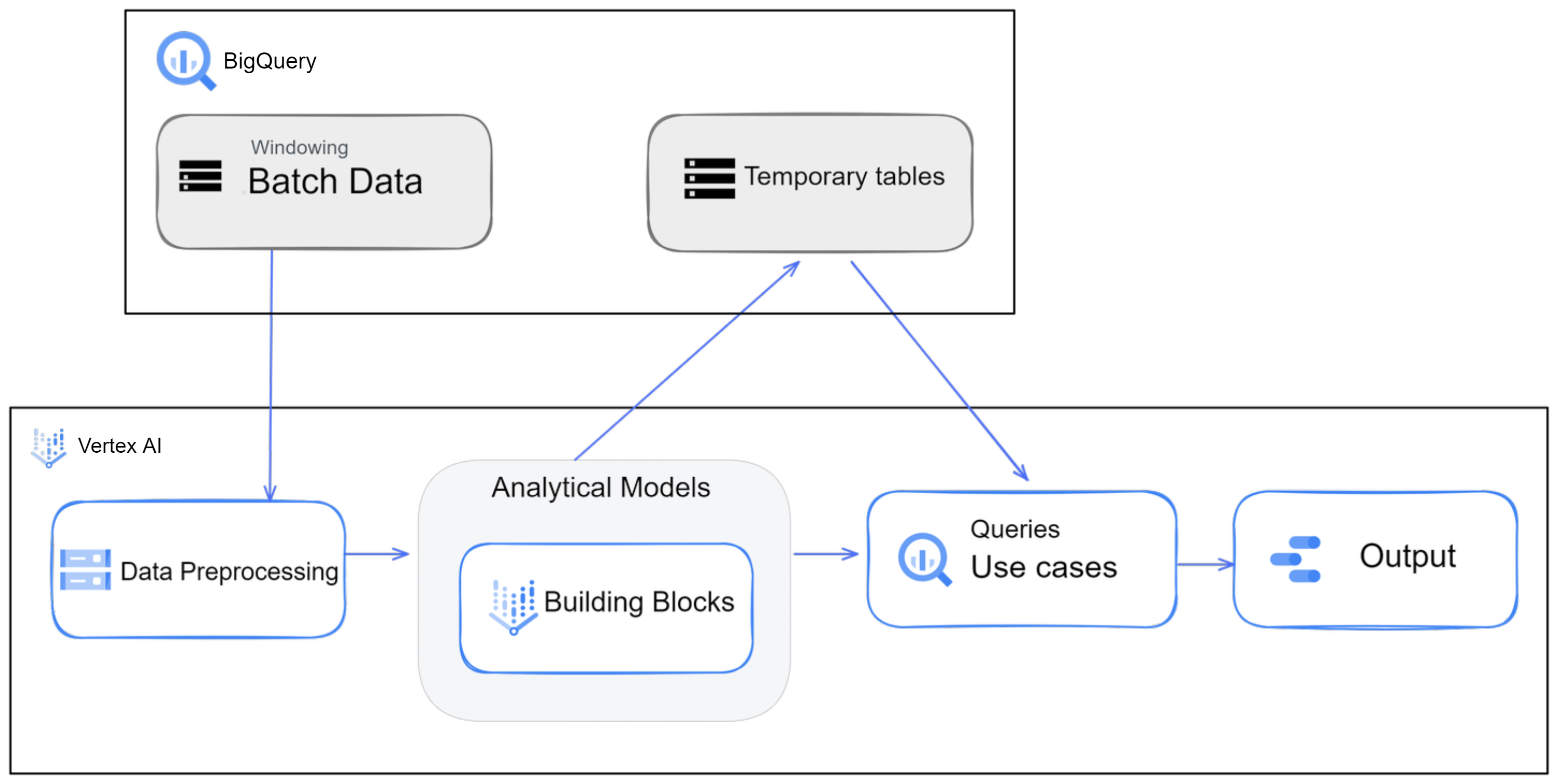}
\caption{Data Analytics Architecture}
\label{fig:architecturedata}
\end{figure}

\section{User trajectories from mobile communications}

\subsection{Definition of trajectory}

Mobile communication data is provided by the mobile phone operator in the form of communication \textbf{logs} between the smartphone of a client in the operator's network and a Radio Access Network (RAN) tower and the location of the client is inferred using proprietary algorithms belonging to the mobile phone operator. Thus, a \textbf{log} is a data point of the form $L =$ (user, latitude, longitude, timestamp) that records the pseudo-anonymized user identifier, the geographic coordinates and time with precision to the nearest second of the request to the communication antenna. Further information is given to inform the client profile, such as the Mobile Country Code (MCC) of the Subscriber Identity Module (SIM) card.

\sloppy
In accordance with the definition of \citep{feng2016survey}, we consider a \textbf{trajectory} to be a sequence of logs of a single user, $Traj = ((u, lat_1, lon_1, T_1)$, $(u, lat_2, lon_2, T_2)$, $\ldots$, $(u, lat_n, lon_n, T_n))$, with the following properties:

\begin{itemize}
    \item $T_i > T_j \iff i > j$.
    \item $T_i - T_{i+1} < \Delta T$, where $\Delta T$ denotes the maximum period between consecutive logs in a trajectory. We define $\Delta T$ to be 30 minutes in this work.
    \item $\delta(lat_i, lon_i, lat_{i+1}, lon_{i+1}) < \Delta D$, where $\delta: \mathbb{R} \times \mathbb{R} \times \mathbb{R} \times \mathbb{R} \rightarrow \mathbb{R}$ is a function that given two geographic coordinates calculates the distance between them and $\Delta D$ is the maximum distance between consecutive logs in a trajectory. We abbreviate this notation to $\delta(P_i, P_j)$ to denote the distance between the coordinates of logs $i$ and $j$ of a given trajectory. As distances between points in our dataset are small enough that the curvature of the Earth can be ignored, we use the Euclidean distance as $\delta$ and define $\Delta D$ to be 1 kilometre. 
\end{itemize}

A trajectory is called \textbf{stationary} if all logs are within a radius of the initial point of the trajectory, i.e., $\delta(P_1, P_i) < r_{SP}, \; \forall i \leq n$. In turn, this means that the maximum distance between any points in a stationary trajectory is $2r_{SP}$. We define $r_{SP}$ to be 200 meters.

A non-stationary trajectory is valid if the following properties are fulfilled:

\begin{itemize}
    \item $\sum\limits^n_{i=2}\delta(P_{i-1},P_i) / (T_n - T_1) < \varepsilon_v$. The average speed of the user cannot exceed a maximum value (35 m/s), as exceeding this value is indicative of errors in data collection.
    \item Let $V_{Traj}$ denote the average speed for the trajectory $Traj$, calculated as above. Then $V_{Traj}/(T_n - T_1) < \varepsilon_a$. The average acceleration also cannot exceed a maximum value (9.8 $m/s^2$), for the same reason.
\end{itemize}

A \textbf{stay point} in a trajectory is a significant segment of the trajectory where the client has little or no movement. Significance is determined by a minimum length of time ($\varepsilon_{SP} =$ 15 minutes) the user must be in a certain radius ($r_{SP}$). Formally, a stay point is a sequence of logs $SP = (L_i, L_j)$, $1 \leq i \leq j \leq n$, such that $\delta(P_i, P_k) < r_{SP}$, $i \leq k \leq j$, and $T_j - T_i > \varepsilon_{SP}$.

\subsection{Data Preprocessment}
\label{datapreprocess}
User communication logs reach the server for analytical processing daily as a big data stream, which we divide into chunks of one hour for ease of processing. Within this hourly data stream, we segment logs to identify trajectories that are \textbf{completed} instead of the remaining incomplete ones that could continue into the next hourly data chunk. This segmentation as a stream technique works by selecting users that have no data logs in the last $\Delta T$ minutes of a given hour or users that contain either a time ($\Delta T$ minutes) or distance ($\Delta D$) gap between consecutive logs and considering the logs solely before this gap.

Upon determining which trajectories are completed, we apply trajectory preprocessing techniques, as suggested by Zheng~\citep{zheng2015trajectory}, namely noise filtering, compression, and stay point detection. For noise filtering, we use both a median filter~\citep{andrade2018identifying} with a window size of 5 and a heuristic-based outlier detection filter~\citep{andrade2020discovering}, with maximum values of instant velocity and acceleration set to 40 $m/s$ and 15 $m/s^2$. It is worth noting that the analysis presented in this work is focused on mobility in urban environments with limited data collected on highways or other high-speed settings, hence the lower thresholds for these values when compared to other values suggested in related work.

Trajectory compression is particularly relevant to our problem, as people often use their mobile or smartphones when standing still, but the data logging process keeps on recording data, and due to the influence of GPS signal loss and data drift, several unwanted points show up. We use an online data reduction technique~\cite{douglas1973algorithms} for compression, in particular the sliding window algorithm~\cite{ramer1972iterative}, discarding data points within a 25-meter radius of the anchor point. We modify the implementation of this algorithm within sci-kit-mobility~\cite {scikit-learn} to preserve the first and last points of the compressed sequence to avoid losing information on length-of-stay when users are at specific locations. This compression method reduced the amount of data stored between 75 and 80\%.

Finally, we apply the stay point detection algorithm of Li et al.~\citep{ye2009mining} to identify stay points with the characteristics detailed in the previous section.

\section{Building blocks}

As previously mentioned, the City Analyzer project involves a consortium of industrial partners from multiple sectors -- tourism, commerce, and transportation. Despite the inherent diversity of these industrial domains, we identify recurring tasks every day to the business requirements of all partners and develop analytical models to solve them, which we refer to as \enquote{building blocks}. In this section, we describe these building blocks.

\subsection{Trajectories crossing polygons}

Our analysis's most fundamental building block is matching trajectories to polygons of interest to know whether a trajectory intersects a geographical area. This building block is helpful to some questions posed by the use case owners, for instance, to know how many clients are in a certain area, how long they stay there or how many times clients visit the area over some time. This building block is also part of the remaining analytical models that require knowing whether a certain user is in a given geographical area. The input to this building block is an area of interest and an interval of time. The output is a list of users, their respective trajectories that cross the polygon and the times of entry and departure into/out of the polygon.

To calculate the time of entry, we search for pairs of consecutive points in a trajectory where a user was first outside the polygon, followed by inside the polygon. Similarly, to calculate the departure time, we search for pairs of consecutive points in the trajectory first inside and then outside the polygon. We also consider that a user's first data point may already be inside the polygon of interest, in which case we mark the time of entry as the time of this first data point and do the same in case a user's last data point is inside the polygon.

\subsection{User Profile}
\label{userprofile}
Profiling is a technique used to identify users' behavioral patterns based on specific properties available in a particular context. In geospatial mobility data, existing literature predominantly focuses on crafting individual user profiles by leveraging features derived from their travel patterns. These features commonly include spatial distribution or visited locations, such as tourist spots or points of interest (POIs). However, our approach diverges by focusing on the locations of stay rather than movement to delineate user profiles. This shift in perspective enables a more nuanced understanding of user behaviour, emphasizing stationary contexts to extract valuable insights into users' places of stay and their residence and occupational statuses. For instance, we take a reverse perspective instead of adopting the common practice of classifying users staying in a hotel as tourists. We focus on identifying users' frequent stay locations and derive relevant features to ascertain their residency status. Subsequently, we examine whether these identified locations coincide with hotels or not.

Here, we discuss the five models we developed for building User profiles, which are also the basis for other building blocks. All the models presented below operate incrementally, processing data daily. The temporal dimension for each day is bifurcated into $daytime$ and $nighttime$ intervals. According to our domain experts, the hours are defined as $08:00 \leq {daytime} < 20:00$ hours and $20:00 \leq {nighttime} < 08:00$ hours of the following day. After each day given the input of user locations (latitude and longitude) with timestamps, the following statistics are incremented for each user \(i\) from the evolving (users getting added or deleted) set of users \(\mathcal{U}\) and area \(a\) from the finite set of areas \(\mathcal{A}\), where \(\mathcal{A}\) can be tessellated into different hierarchical levels of geography such as S2 cells, subsections, sections, parish, municipality, district and so on.
\begin{itemize}
    \item \(night\_count_i^a\): Number of nights spent by user \(i\) in the area \(a\)
    \item \(day\_count_i^a\): Number of days spent by user \(i\) in the area \(a\)
    \item \(stay\_time\_day_i^a\): mean time spent in the daytime by user \(i\) in the area \(a\)
    \item \(stay\_time\_night_i^a\): mean time spent in the nighttime by user \(i\) in the area \(a\)
    \item \(length\_of\_stay_i^a\): Duration of stay (nights $\cup$ days) spent by user \(i\) in the area \(a\)
\end{itemize}
We omit explicit mention of individual users ($u$) and areas ($a$) for all subsequent references to simplify notation.

\subsubsection{User Night Stay Location Identification}
We define a user's night stay location as the primary place where they spend most of their time during the $nighttime$, often synonymous with their home. The S2 cell corresponding to this night stay location is identified in this context. This determination is made within the framework of a given geographical area, which is tessellated into S2 cells. For each S2cell $a$ and user $u$ we calculate the statistics \(nightcount_i^a\), \(daycount_i^a\), and \(staytimenight_i^a\). The night stay place is then labelled as {\tt "yes"} or {\tt "no"} based on the implemented rules:
\begin{itemize}
    \item $night\_count > 0$
    \item $stay\_time\_night > \theta$
    \item The location with the highest $stay\_time\_night$ meeting these criteria is classified as the night stay place for the user.
\end{itemize}

\subsubsection{User Day Stay Location Identification}
The definition of a user's daytime stay location mirrors that of the nighttime stay location, with the distinction being the focus on daytime activities. This characterization is pivotal in identifying a user's work location. The corresponding S2 cell associated with the daytime stay location is pinpointed within a given geographical area, which undergoes tessellation into S2 cells for structured analysis.
Similarly, for each S2cell $a$ and user $u$ we calculate the statistics \(daycount_i^a\), \(nightcount_i^a\), and \(staytimeday_i^a\). The night stay place is then labelled as "yes" or "no" based on the implemented rules:
\begin{itemize}
    \item $day\_count > 0$
    \item $stay\_time\_day > \theta$
    \item The location with the highest $stay\_time\_day$ meeting these criteria is classified as the day stay place for the user.
\end{itemize}

\subsubsection{Residency Status Classification}
\label{Residencystatusclassification}
This is one of the pivotal blocks in the framework and serves as a basis for many other blocks. A user's residency status is labelled as Resident, Tourist or Visitor. A user's residency status is calculated for different levels of hierarchy, basically S2cell, section, parish, municipality and district level. The user location data (latitude, longitude), timestamp and MCC code are inputted. For each user and the area in which all the features defined above \(day\_count_i^a\), \(night\_count_i^a\), \(stay\_time\_day_i^a\), \(stay\_time\_night_i^a\), \(length\_of\_stay_i^a\) are updated per day by merging the previous day information with the current statistics. The labels of users are updated for every geography in the hierarchy based on the following conditions outlined in the algorithm below:

\begin{algorithm}
\scriptsize 
\caption{Determining Residency Status}
\begin{algorithmic}[1]
\Require{$\text{A row of user data with attributes:}$ $\textit{old\_res\_status, new\_res\_status, }$ $\textit{ length\_of\_stay, night\_count, stay\_time\_night, 
stay\_time\_day, mcc}$}
\Ensure{Residential status of the user as 'Casual Visitor', 'Regular Visitor', 'Commuter', 'National Tourist', 'International Tourist', or 'Resident'}
\Function{DetermineResidentialStatus}{row}
    % Rule 1
    \If {$residential\_status\_x$ is null}
        \State \Return $residential\_status\_y$
    \EndIf

    % Rule 2
    \If {$residential\_status\_y$ is null}
        \If {$length\_of\_stay < \text{min\_length\_of\_stay}$ and $stay\_time\_night < \text{min\_night\_stay\_time}$ and $stay\_time\_day \geq \text{max\_commute\_time}$}
            \State \Return 'casual visitor'
        \ElsIf {$length\_of\_stay \geq \text{min\_length\_of\_stay}$ and $stay\_time\_night < \text{min\_night\_stay\_time}$ and $stay\_time\_day \geq \text{max\_commute\_time}$}
            \State \Return 'regular visitor'
        \Else
            \State \Return $residential\_status\_x$
        \EndIf
    \EndIf

    % Rule 3
    \If {$stay\_time\_night < \text{min\_night\_stay\_time}$ and $stay\_time\_day < \text{min\_night\_stay\_time}$}
        \State \Return 'commuter'
    \EndIf

    % Rule 4
    \If {$length\_of\_stay < \text{min\_length\_of\_stay}$ and $\text{min\_night\_stay\_time} \leq stay\_time\_night < \text{min\_night\_stay\_time}$ and $stay\_time\_day < \text{max\_commute\_time}$}
        \State \Return 'casual visitor'
    \EndIf

    % Rule 5
    \If {$length\_of\_stay \geq \text{min\_length\_of\_stay}$ and $\text{min\_night\_stay\_time} \leq stay\_time\_night < \text{min\_night\_stay\_time}$ and $stay\_time\_day < \text{max\_commute\_time}$}
        \State \Return 'regular visitor'
    \EndIf

    % Rule 6
    \If {$length\_of\_stay \geq \text{min\_length\_of\_stay}$ and $night\_count \geq 0.0$ and $stay\_time\_day \geq \text{max\_commute\_time}$ and $stay\_time\_night < \text{min\_night\_stay\_time}$}
        \State \Return 'regular visitor'
    \EndIf

    % Rule 7
    \If {$length\_of\_stay < \text{min\_length\_of\_stay}$ and $night\_count \geq 0.0$ and $stay\_time\_day \geq \text{max\_commute\_time}$ and $stay\_time\_night < \text{min\_night\_stay\_time}$}
        \State \Return 'casual visitor'
    \EndIf

    % Rule 8
    \If {$night\_count > 0.0$ and $night\_count < \text{max\_night\_count}$ and $stay\_time\_night \geq \text{min\_night\_stay\_time}$}
        \If {$mcc == '268'$}
            \State \Return 'national tourist'
        \Else
            \State \Return 'international tourist'
        \EndIf
    \EndIf

    % Rule 9
    \If {$night\_count \geq \text{max\_night\_count}$ and $stay\_time\_night \geq \text{min\_night\_stay\_time}$}
        \State \Return 'resident'
    \EndIf
\EndFunction
\end{algorithmic}
\end{algorithm}

\subsubsection{Residency area classification}

The objective of this component is to classify the location where a user spends a certain amount of time, categorizing it as either their home, a hotel, undefined, or other type of location. We utilize information about amenities in a given geographical area to achieve this goal, specifically focusing on hotels. With access to this data, we employ map matching techniques to correlate user night-stay locations obtained from building block 3 with buffered* hotel areas. By doing so, we can determine whether a user stays in their residence, a hotel, or another (neither their home nor a hotel). In cases where there is insufficient information to ascertain the user's night-stay location, we label it as "undefined."

\subsubsection{Professional status classification}
In line with the preceding block, which used external data sources to furnish information regarding amenities within a locale, we extend our approach to incorporate data concerning the geographical boundaries of educational institutions. This data is then used to correlate with the user's daytime stay location, resulting from Building Block 2.2 (implicitly assuming these to be places of employment, excluding those who operate from home). The coordinates of daily stays are mapped against the buffered zones encompassing educational institutions. Those intersecting with such zones are categorized as 'students'. Individuals whose daily stays remain unidentified are designated as 'other', while the remainder are classified as 'workers'."

\subsection{Mode of transport classification}

Determining the mode of transport people use to move between locations is an important facet of urban mobility analysis, allowing outcomes such as public transportation route planning. This importance has led to the emergence of a rich methodological field aimed at automatically identifying the mode of transport given a trajectory~\cite{andrade2022you}. However, most work in this area relies on GPS data, which is more regular and precise than location inferred from mobile communications data. In addition to the challenges inherent to this task, our dataset is completely unlabeled, limiting the number of previous approaches applicable to our context.

Considering the challenges we have just outlined, we propose a semi-supervised learning model based on pseudo-labeling~\cite{cascante2021curriculum} and co-training~\cite{chen2022semi} for a mode of transport identification. Our approach works by considering two classifiers pre-trained on the popular trajectory mining dataset GeoLife~\cite{zheng2010geolife} and iteratively refining them until an agreement threshold is reached. See Figure~\ref{fig:SSL_MOT} for a diagram of the models and training scheme, more details follow.

\begin{figure}
    \centering
    \caption{Semi-supervised framework for mode of transport classification, based on pseudo-labeling and co-training.}
    \label{fig:SSL_MOT}
\end{figure}

The GeoLife dataset \citep{zheng2010geolife} is a dataset collected by Microsoft Research Asia, tracking the GPS location of a set of people over a long period. The most relevant aspect of this dataset for our development is that some trajectories are labelled with the mode of transportation, including different labels for different legs of a single journey (for example, someone using the train and reaching their final destination by foot). The available labels include train, taxi, walk, bus, subway, aeroplane, car, bike, boat, run and motorcycle. We drop trajectories labelled aeroplanes, boats, and motorcycles, merge trains with subways, cars with taxis, and walk with a run. After removing trajectories with less than 10 data points, the final dataset contained 8191 trajectories split among the five modes of transportation labels: 3416 on foot, 1692 by bus, 1250 by car, 1121 by bike and 712 by train. To become less susceptible to class imbalance issues, we randomly undersample the \textit{on foot} class so that the training data has 2000 examples belonging to this class.

To train our mode of transport classifier, we start by training an initial classifier on the GeoLife dataset based on the work of \cite{andrade2022you}, who concluded that a Random Forest Classifier (RF) trained on trajectories described by their velocity and acceleration aggregated statistics (average, median, standard deviation and maximum value) can separate mode of transportation classes without external data. Our co-model for the co-training learning scheme is a Multilayer Perceptron (MLP), which sees a different view of the data by a normalization pre-processing step applied to the features. We used the scikit-learn \citep{scikit-learn} implementation for both these models, tuning the maximum depth of each decision tree and the number of estimators for the RF and both the number of layers and the number of neurons per layer for the MLP using 10-fold cross-validation, keeping the rest of the hyperparameters equal to the default.

Upon training the two initial models, we transfer them into the domain of our mobile communications data and start the pseudo-labeling phase. We perform inference on the trajectories extracted from the mobile communications data and assume as true that the trajectories are assigned the same label by the two initial models. We gather this subset of trajectories as new training data and re-train an RF and an MLP using the same procedure (hyperparameter tuning with cross-validation). This process is repeated until the two models agree on 5\% of the labels, after which the RF model is saved and used for inference.

\subsection{Origin-destination - Traffic flows}

We approach the problem of road traffic analysis from an anomaly detection perspective, attempting to find a city's normal spatial and temporal origin-destination traffic flow and creating a metric that highlights when the traffic flow deviates from normal behaviour. This approach is grounded on the requirements set out by this project, in contrast to related literature on traffic analysis that tackles the issue from a variety of approaches, allowing us to answer questions such as what are the optimal locations for advertisement or when should mobile network capacity be boosted to meet mobile traffic requirements that deviate from normal.

Our methodology is based on a tessellation of an input area, that is, dividing the input geography into smaller units that cover the whole initial area to map continuous geographical coordinates and the discrete regions. The anomaly detection module is agnostic to the tessellation of choice, as the tessellation method should vary according to the end application of the method. For example, an application that requires knowledge of traffic in a fine-grained area may consider a tessellation based on Google S2 cells; on the other hand, if such fine-grained detail is unnecessary or uninformative, we have also considered tessellations based on municipality and parish borders (Nomenclature of Territorial Units for Statistics (NUTS) levels 4 and 5).

The metric used to identify temporal anomalies is based on the traffic time series for a given unit of the chosen tessellation over one week. Given a tessellation with k units $\mathcal{T} = \left\{ C_1, C_2, \ldots, C_k \right\}$, let $\zeta_i^t$ denote the traffic in cell $C_i$ at time $t$, where $t$ denotes one hour (for example, from 00:00 to 01:00). Then, we define the temporal anomaly score ($TAS$) as:

\begin{equation}
    Z_i^t = \frac{\zeta_i^t - \overline{\zeta_i}}{\sigma_{\zeta_i}}, \qquad TAS_i^t = \frac{Z_i^t}{\sqrt{ \sum\limits_k Z_i^k}},
\end{equation}
where $\overline{\zeta_i}$ is the mean traffic in cell $C_i$ over the week before time $t$ (previous 168 hours) and $\sum\limits_k Z_i^k$ indicates the sum of $Z_i$ over the same time period. Thus, $Z_i^t$ is the Z-scored traffic of a cell concerning its traffic over the previous week, and $TAS$ is simply the normalized Z-score. Similarly, we define the spatial anomaly score ($SAS$) as the normalized Z-score for a certain time across all cells of a given tessellation:

\begin{equation}
    {Z'_i}^t = \frac{\zeta_i^t - \overline{\zeta^t}}{\sigma_{\zeta^t}}, \qquad SAS_i^t = \frac{{Z'_i}^t}{\sqrt{ \sum\limits_k {Z'_k}^t}},
\end{equation},
where $\overline{\zeta^t}$ is the mean traffic across the area of study at time $t$ and $\sum\limits_k {Z'_k}^t$ indicates the sum of $Z^t$ over the same period

\subsection{Frequent trajectories}

With the development of location-based positioning devices and the advent of the Internet-of-Things (IoT), more and more moving objects are traced, and their trajectories are recorded, joining diversified information about their carriers and equipment. These data comprise a rich source of spatial and temporal semantic information. Therefore, moving object trajectory clustering undoubtedly becomes the focus of the study in moving object data mining \citep{zheng2015trajectory}.

Many areas can leverage the similarity of trajectory analysis, such as policy-makers/government, transportation companies/authorities, last-mile parcel carriers, biologists, and retail and marketing companies. In the public sector, the managers can analyze the moves of people at different hours of the day and week to promote changes in the infrastructure of a region, change the bus routes, increase the number of trains or metro cars, and take measures to diminish the bottlenecks of traffic hot-spots and try to diminish the vehicle emissions. The private sector can also use these studies to target advertisements to specific groups of users that travel along some routes or visit some points of interest. Biologists can use these techniques to help understand the whereabouts of animals such as birds and fish, where they go, and which are the recurrent routes taken. Tourism in both public and private sectors can also make good use of trajectory analysis by recommending tourist routes or using these routes to improve or even deploy services along the path.

A common approach to performing trajectory analysis is by making use of clustering techniques where the process assigns a set of similar trajectories into groups (the clusters) having highly similar trajectories within each cluster and low similar trajectories among the different cluster sets \citep{zheng2015trajectory, yuan2017review}. Among the clustering approaches, one, in particular, has shown to be more suitable for trajectory analysis due to its possibility of forming clusters of arbitrary shapes in Euclidean space: the density-based approach. One of the most popular algorithms in this group is DBSCAN \citep{ester1996density}. Still, one key component of good-quality trajectory analysis is how to calculate the similarity between trajectories in a group. Different similarity measures can be used, but not all consider the order of the data points in the trajectory set, which is paramount for a good-quality cluster of similar trajectories. Fréchet distance is one of the metrics that can be used to solve the problem.

\textbf{Frequent Trajectory:} A frequent trajectory is described as a regular route an individual tends to follow when travelling/moving between two locations (origin and destination) \citep{andrade2020discovering}. A real-life example can be a street or highway to drive from home to work, a metro line used to commute, a sidewalk to walk to the user's preferred restaurant, a shopping mall, etc. In this study, we have focused on discovering the most frequent places that individuals visit and the common routes related to these displacements.

Other characteristics are also important to mention: 
\begin{itemize}
    \item Trajectories may have different lengths as individuals tend to move accordingly to their needs and singularities (e.g., \textit{N} and \textit{M} can be different for \textit{Tr}$_i$ \texttt{=} $(p_1, p_2, $\ldots$, p_N)$ and \textit{Tr}$_j$ \texttt{=} $(p_1, p_2, $\ldots$, p_M)$). 
    \item Trajectories may have different directions. In the context of individuals' movement, the direction of each trajectory is an essential condition for the similarity of trajectories. As we propose the discovery of frequent routes, two trajectories moving in opposite directions should be considered different moves despite their proximity. They may represent different habits (e.g., going to work from home and going back home from work).
\end{itemize}

\subsubsection{Trajectories Simplification with Ramer-Douglas-Peucker (RDP)}

In some cases, GPS raw data can be very densely represented. The three datasets used in this work have different granularities, and we do not need much detail for the frequent trajectories discovery/clustering step. Many of these points can be removed as they are somehow redundant, whereas other key positions must be kept. An excellent way to avoid unnecessary processing is by using compression techniques. We use the Ramer-Douglas-Peucker algorithm \citep{ramer1972iterative, douglas1973algorithms} to simplify the trajectories. The algorithm aims to produce a simplified poly-line with fewer points than the original but still keeps the original’s characteristics/shape. The method takes one threshold parameter $\epsilon$ and connects the original line's first and last point with a reference OD pair. It then finds the point furthest away from that baseline reference and checks if it's greater than $\epsilon$. If true, it keeps the point, and the function recursively splits the line into two segments, creating new reference points and repeating the procedure. If the point is nearer to the baseline reference than $\epsilon$ it discards all the points between these reference points simplifying the trajectory. 

Figure \ref{fig:rdp_frechet} (a) shows an example of a trajectory split by the RDP algorithm.

\subsubsection{Clustering Algorithm and Similarity Measures}

Clustering is an efficient way to group data into different classes based on the internal and previously unknown schemes inherent in the data, and trajectory clustering is the most popular topic in current trajectory data mining. The aim is to discover the similarity (distance) in moving object databases, grouping similar trajectories into the same cluster, and finding the most common patterns \citep{yuan2017review}.

Density-based clustering techniques are very popular methods for location detection. They can detect clusters of arbitrary shapes without specifying the number of clusters in the data beforehand. Furthermore, they are tolerant of outliers (noise). 
Some recent studies have addressed the location detection techniques to improve the quality of the discoveries \citep{andrade2018identifying,andrade2019discovering,andrade2020mining,andrade2020mobility,andrade2020discovering}. 

In this study, we apply the clustering method proposed by \citep{andrade2020discovering}, which is a variation of DBSCAN \citep{ester1996density} to form the clusters of trajectories between the start (origin) and end (destination) points of all the trajectories. 

One of the most important parts of a clustering algorithm is the similarity measure of two items. This is the step where the distance of two points is calculated before the algorithm decides whether to group these items. Different comparison strategies must be taken according to the purpose of the clustering task. Some of the most common distances are Euclidean, Hausdorff \citep{zheng2015trajectory}, Longest Common Sub-Sequence \citep{vlachos2002DSM}, Dynamic Time Warping \citep{zheng2015trajectory}, and Fréchet distance \citep{eiter1994computing}.

For the Euclidean distance (ED), the similarity between the two trajectories is simple and intuitive because it is parameter-free. In addition, its time complexity is linear, meaning it can handle a large dataset. However, noise existing in trajectory data will have a great influence on the result. Another main disadvantage of using the ED for measuring the similarity between trajectories is that the sampling points must be in corresponding positions (at the same time), and the trajectories must have the same length. This is not true in real-world scenarios, even though the origin and destination are the same.

Hausdorff distance (HD) between trajectory segments A and B selects the maximum unidirectional HD from A to B and from B to A. It measures the maximum mismatching degree between two trajectory segments. HD tolerates the influence caused by point disturbance but is sensitive to noise data. This last point is also an issue in real-world scenarios when dealing with GPS data due to the signal interference caused by objects. For this reason, we avoid using this distance function.

For the Longest Common Sub-Sequence (LCSS), as the name suggests, the idea is to get the longest list of common items in sequence between two trajectories. It uses a distance function (ED or any other) to compare if the combination of a pair of points is less than a threshold $\epsilon$. Having the distance value less than the expected threshold, the value of LCSS is increased by 1. One advantage of LCSS is that it allows certain deviations in the sampling data (common in the real world). The advantages are the distance measure choice and parameter specification as well as the time complexity.

The Dynamic Time Warping (DTW) algorithm was proposed to find an optimal alignment between two given (time-dependent) sequences under certain restrictions. This method can match trajectories even if their lengths are different. The goal is to minimize the warping cost of finding similar paths between two trajectories. It is also sensitive to noise. The disadvantages are that when two trajectories are completely dissimilar in a small range, the DTW distance cannot be found, and the time cost and complexity are higher than the previous techniques.

\begin{figure}[!h]
\centering
    \begin{minipage}[]{0.49\linewidth}
        \centering
        \includegraphics[width=1\textwidth]{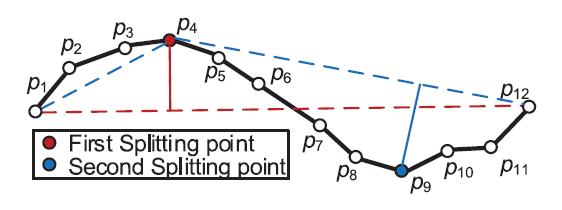} \\
        (a) Douglas-Peucker algorithm \citep{zheng2015trajectory}
    \end{minipage}
    \begin{minipage}[]{0.49\linewidth}
        \centering
        \includegraphics[width=1\textwidth]{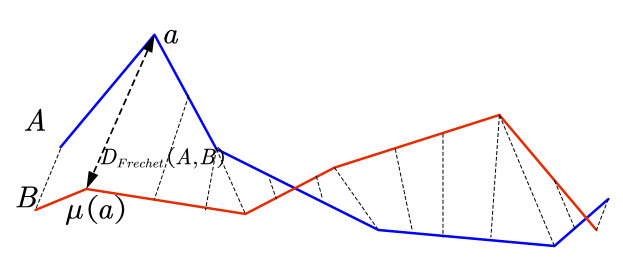} \\
        (b) Fréchet distance between trajectories
    \end{minipage}
\caption{RDP and Fréchet distance examples} \label{fig:rdp_frechet}
\end{figure}

Finally, the discrete Fréchet Distance (DFD) considers the sequential relationship and the location of the points in the trajectories while measuring their similarity. It also relies on ED to calculate the distance in a point-wise fashion, as shown in equation \ref{eq_DFD}.

\begin{equation} 
DFD(x, y) = \max \big(\|p_{i(t)} - q_{i(t)}\|, \min {\big( DFD(x-1, y), DFD(x, y-1) \big)}{\big)} \label{eq_DFD}
\end{equation}

where given two sequences of points $p$\textit{=}($p_1$, $p_2$, $p_3$, \ldots, $p_n$) and $q$ \textit{=} ($q_1$, $q_2$, $q_3$, \ldots, $q_m$), Fréchet distance represented by $DFD(x,y)$ is the maximum of the minimum distances between points $p_i$ and $q_i$. Figure \ref{fig:rdp_frechet} (b)  shows an example of the Fréchet distance between two trajectories.

One of the critical issues previous studies have shown is that the Fréchet distance contains the temporal relationship between the points. More particularly, the structure of the nodes inside the trajectory is considered in the computation process, which can more accurately describe the similarity between the trajectories, yelling better results \citep{magdy2015review}. In some scenarios with a backward direction, ring, or crisscross in a trajectory, the Fréchet distance value doesn't show more distortion than other distance measurements. Due to these characteristics, this metric is more descriptive and more suitable for measuring the similarity between trajectories. The time complexity is also similar to the other mentioned metrics.

For this work, we used the DFD to cluster the trajectories. We first construct a symmetric distance matrix of each pair in the trajectories connecting the given OxD using the DFD to obtain the clusters of frequent trajectories between the origin and destination pairs. We then fit the symmetric distance matrix to the DBMeans \citep{andrade2020discovering} method to obtain the different groups of trajectories.
\begin{figure}
    \centering
    \includegraphics[width=1\textwidth]{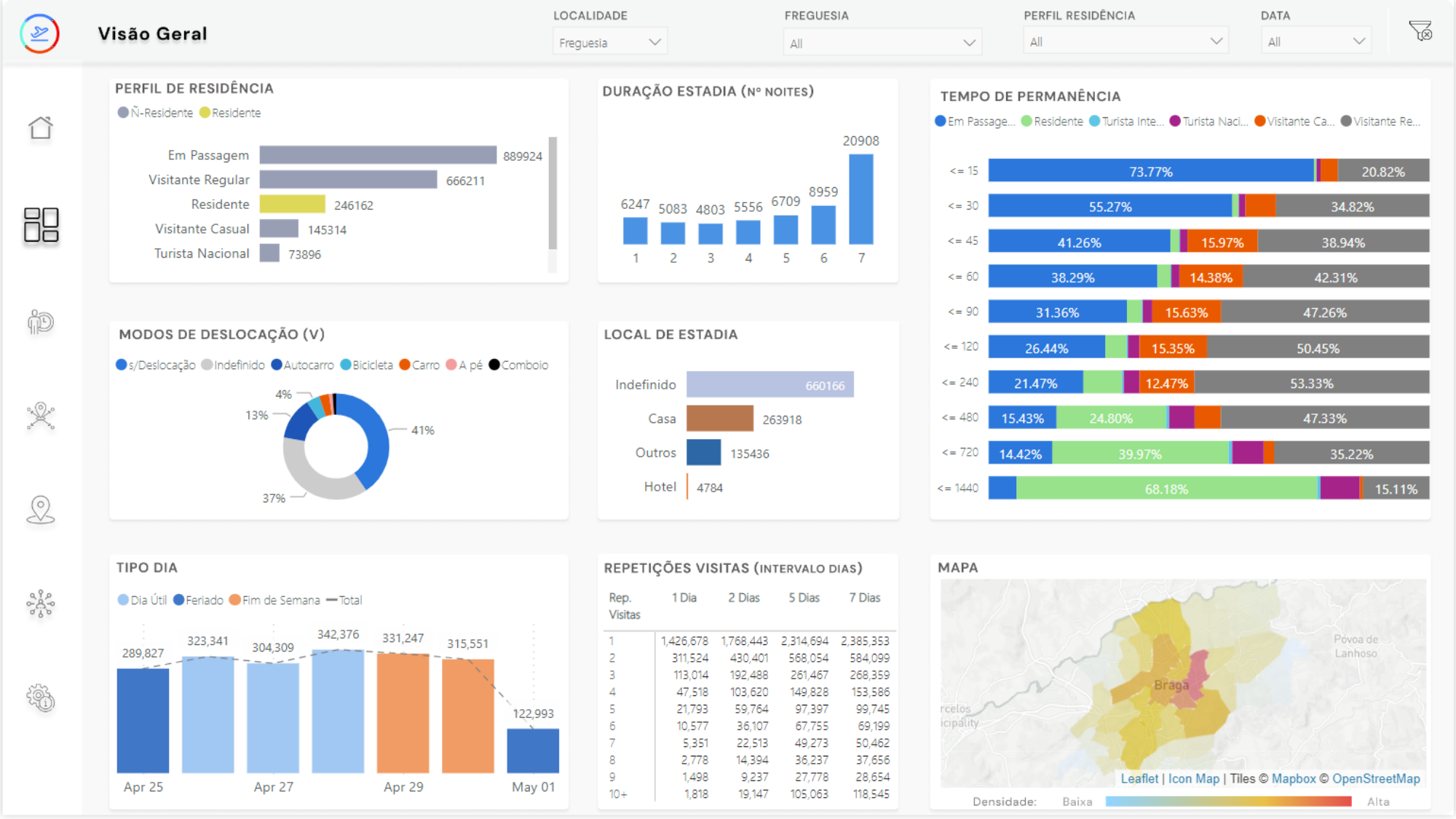}
    \caption{Tourism analytics dashboard for interactive spatial, temporal, and demographic analyses of users' mobility patterns}
    \label{fig:tourism1}
\end{figure}

\section{Use cases}
Below we delineate the use cases identified by stakeholders to guide the utilization of project resources and generate expected outputs. The building blocks discussed above were structured to be shared across multiple use cases. These use cases leverage the models by calling them directly or querying results stored in temporary tables (saved by models) within BigQuery, which continuously updates in real-time against incoming data streams. All the use cases utilize the output table generated by the residency status classification model to categorize and organize the results according to user profiles and other models. Queries for each use case are executed to produce outputs tailored to the needs defined by our partners. These results are generated for specific time intervals and geographic areas, which users can customize as parameters.

\begin{figure}
    \centering
    \includegraphics[width=1\textwidth]{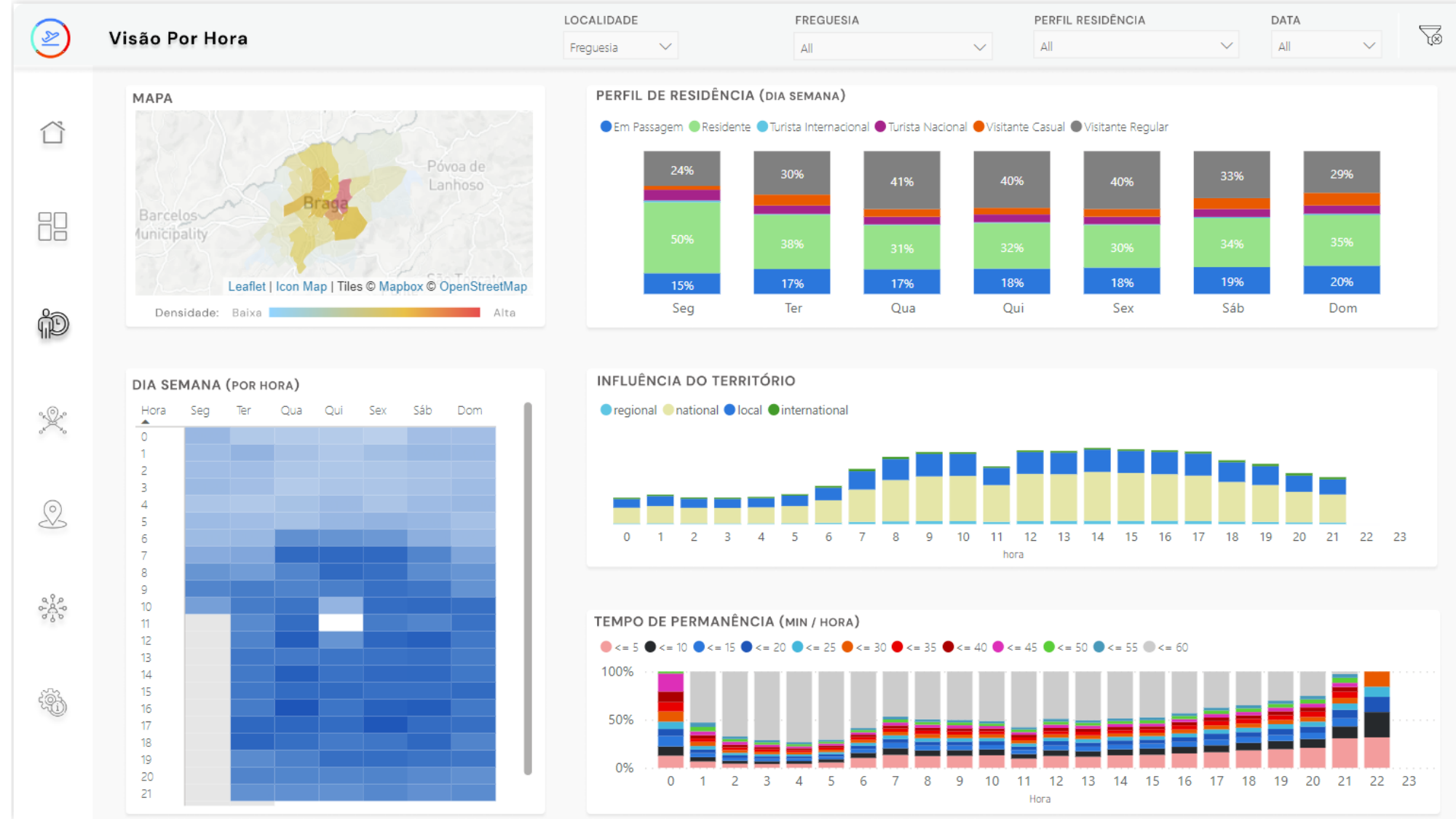}
    \caption{Hourly tourism analytics dashboard to visualize temporal variations in visitor distribution, residency profiles, territorial influence, stay duration, and spatial density, enabling the analysis of hourly mobility patterns.}
    \label{fig:tourism2}
\end{figure}

\begin{figure}
    \centering
    \includegraphics[width=1\textwidth]{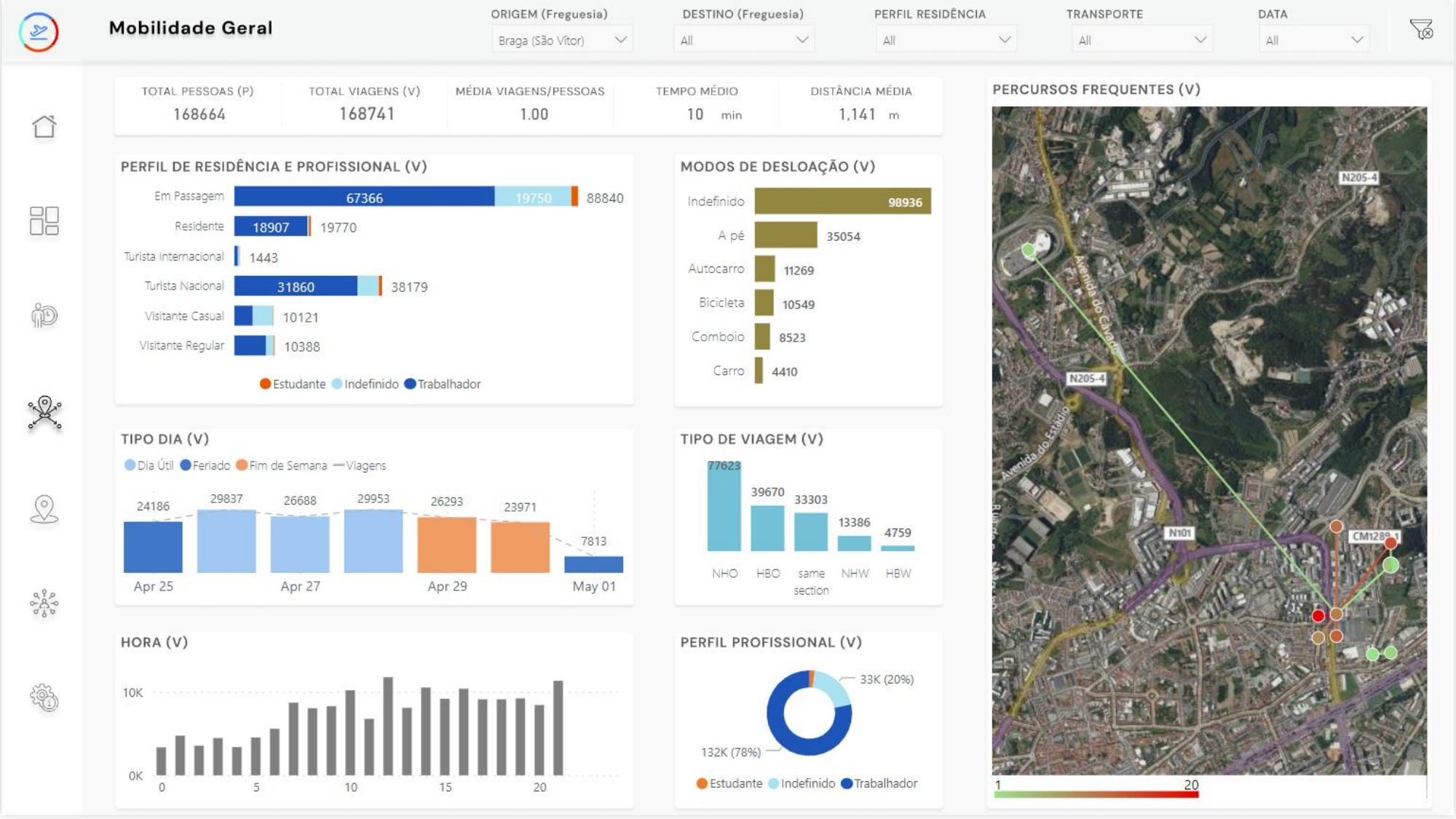}
    \caption{Mobility analytics dashboard to visualize origin–destination patterns, frequent trajectories, transport modes, trip characteristics, and user profiles, supporting transportation planning and mobility management}
    \label{fig:tourism3}
\end{figure}

\begin{figure}
    \centering
    \includegraphics[width=1\textwidth]{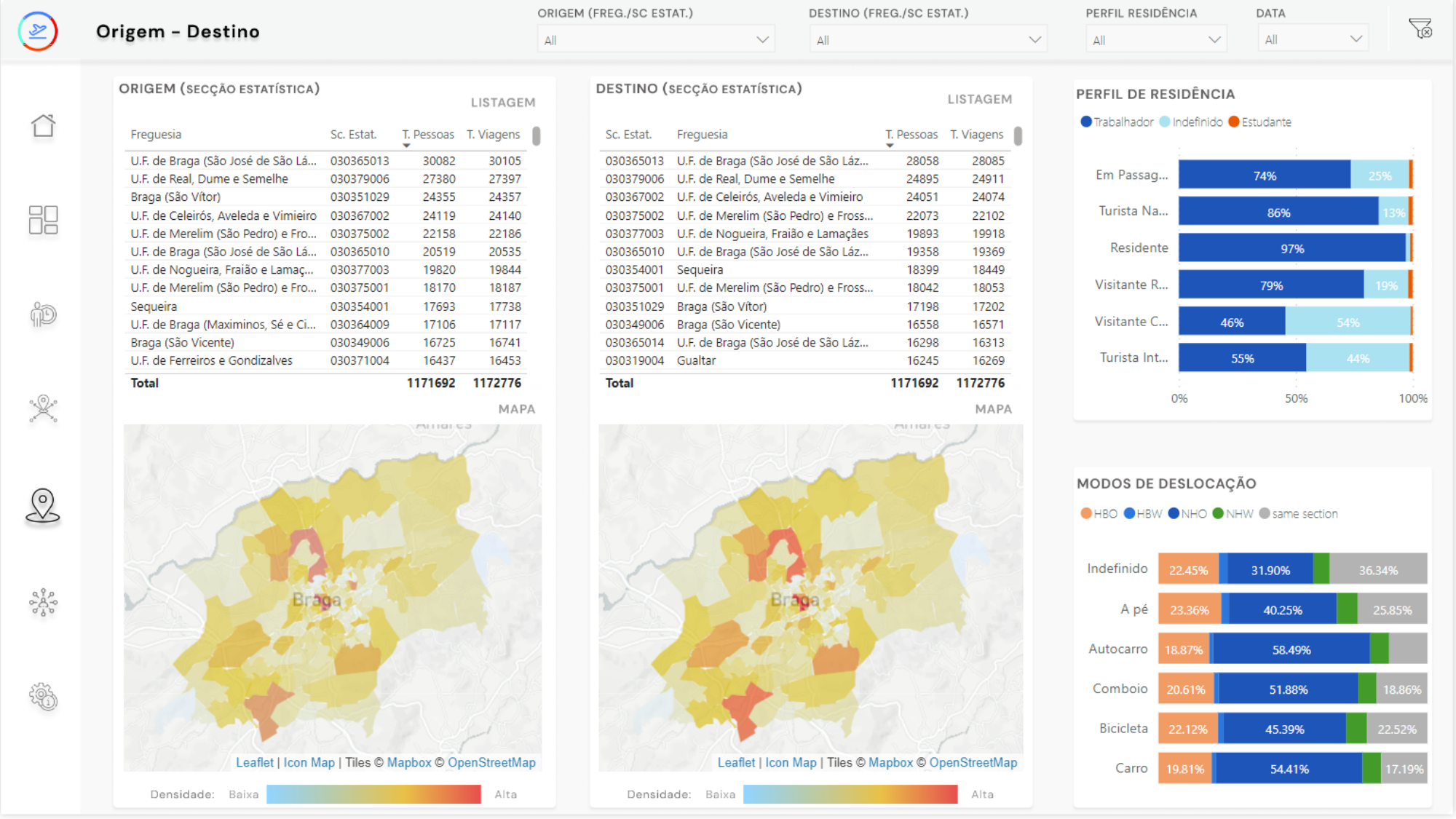}
    \caption{Origin–destination analytics dashboard to visualize trip flows, origin and destination distributions, residency profiles, travel purposes, and mobility statistics, supporting transportation planning and mobility management.}
    \label{fig:tourism4}
\end{figure}

\begin{figure}
    \centering
    \includegraphics[width=1\textwidth]{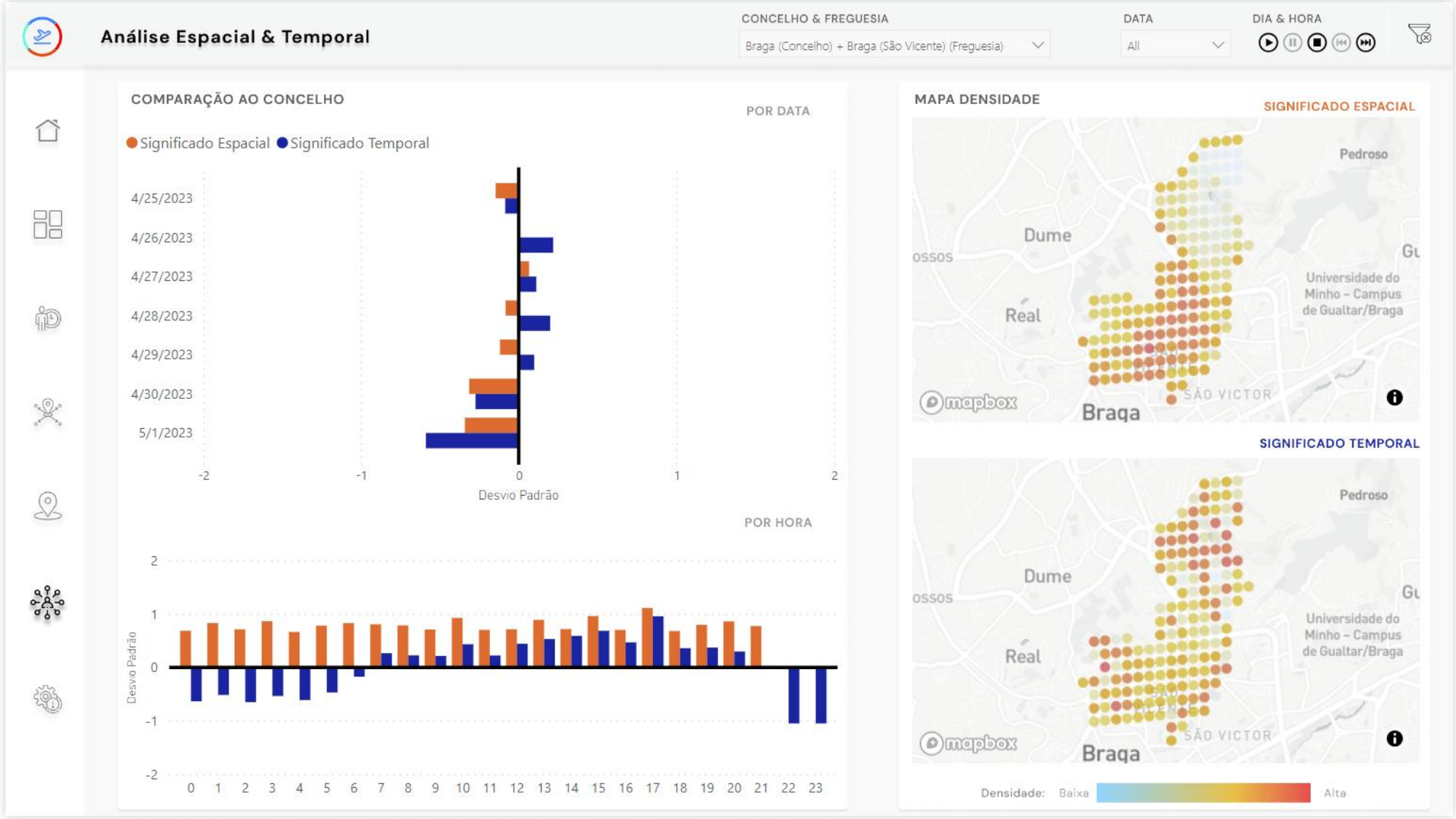}
    \caption{Spatial and temporal anomaly analysis dashboard presenting significance scores and density maps to identify deviations in traffic flow across different geographic areas, dates, and hours, supporting the detection of unusual mobility patterns.}
    \label{fig:tourism5}
\end{figure}

The outputs, serving as data products aligned with the identified use cases, were stored in the cloud or designated buckets. Subsequently, the results were seamlessly transferred to Power BI to create comprehensive reports and dedicated interactive dashboards, enabling different stakeholders to visualize and interpret the results effectively.  Representative dashboards are shown in Figures \ref{fig:tourism1} - \ref{fig:tourism5}. Stakeholders were granted access only to their respective dashboards, which provided a unified view of the multiple mobility indicators relevant to their business requirements. For example dashboard for tourism analysis in Figure \ref{fig:tourism1} integrates multiple data products from the use cases below; including area population counts, time spent, different visits, night-stay classification, stay-location analysis.

\subsection{Area analysis}
\subsubsection{Area Population Count}
This use case involves analyzing the population count within specific geographical areas over hourly or daily intervals while considering the distribution of people from different residency statuses. The geographical areas, represented by polygons such as Municipality, Parish, Section, Subsection, or s2cells, are defined by their unique polygon IDs. The analysis focuses on understanding the distribution and fluctuations of population counts within these areas over time.

\subsubsection{Time spent Analysis}
It aims to analyze the duration of time spent by individuals of different residency statuses within designated geographical areas/polygons. The analysis was conducted hourly and daily at various intervals, such as 5, 10, 15 minutes, and so on. It provides insights on how long the individuals stay in different locations and in which areas people tend to spend more time.
\subsubsection{Different visits}
This use case aims to track the frequency of visits by individuals to particular geographic regions/polygons of interest over a defined time frame, considering their residency status. This involves counting the number of repeat visits each person makes within the designated area.
\subsection{Catchment area analysis}
The catchment area represents the region or territory people travel to access a particular service, facility, or business. In this use case, we analyze the number of visitors in different categories from residency status (international tourists, national tourists, casual visitors, and regular visitors) within specific geographic areas (such as municipalities, parishes, sections, or subsections) during each hour. The analysis distinguishes individuals as local, regional, national, or international. Where "local" and "regional" means the users are residents in the same municipality or district as the geographic area being analyzed. Conversely, if the label is "national" or "international," it indicates that the visitors come from a district or country other than Portugal. This use case provides valuable insights into visitor demographics and their distribution across locations.

\subsection{Night stay analysis}
\subsubsection{Night stay classification}
This use case focuses on understanding where people stay overnight within specific geographic areas during a given period. It tracks the number of individuals staying in different accommodations, such as hotels, homes, or other establishments, based on their night stay location and residency status given by models in section 3. The analysis is conducted within polygons representing various geographical divisions like counties, parishes, sections, or points of interest (POIs).
\subsubsection{Understanding night stay patterns}
The problem involves quantifying the number of individuals who spend nights within various geographic areas (S2 cells, subsections, sections, parish, municipality, district) over specific time intervals classified by their residency status. For instance, within a municipality, the aim is to determine the distribution of residents, tourists, visitors, etc., among the individuals staying there overnight.

\subsection{Stay locations analysis}
This use case involves analyzing the locations where people stay based on their professional status (student, worker, or other) and residency status. It focuses on counting the number of individuals either studying or working within specific geographic areas during a specified period between the start and end dates and their residency status in that area. This information was intended to help transportation planners tailor bus routes and schedules to better meet the needs of students and employees. By optimizing transportation services, the solution aims to improve overall efficiency, reduce congestion, and enhance the commuting experience for residents or visitors.

\subsection{Traffic flow Analysis}
\subsubsection{Hourly Traffic flow}
Here, we analyze the traffic flows hourly within specific geographic cells. For each hour and cell, the number of people present is recorded. Additionally, the significance\_spatial metric compares the number of people in each cell against the average for all cells, indicating over- or under-representation. Similarly, the significance\_temporal metric compares the count of people in a cell over time, providing insights into temporal trends in traffic flow. This analysis helps understand patterns of human activity and movement within different geographic areas over time.
\subsubsection{Origin-destination patterns}
This use case involves analyzing travel patterns between different locations. For each hour, we track the number of trips and unique individuals traveling between specific areas, identified by their origin and destination. The analysis includes details such as the mode of transportation, the residency status of travelers, and the purpose of each trip. Additionally, we provide insights into the average duration and distance traveled for each trip category.
\subsection{Market share of time}
This use case involves analyzing the percentage of time individuals spend inside a specific business/service area compared to other business areas within a defined geographic area (e.g., Municipality, Parish, Section). The analysis is conducted hourly to understand the distribution of time individuals spend across different supermarkets in the area.

\subsection{Frequent trajectories}
The context where the frequent trajectories were applied in the project is related to a public transportation agency where the player wanted to know the mobility profiles and the most common routes used to commute in a given region. The main aim is to optimize service delivery, adapt offers, and identify users' necessities.

%\subsubsection{Data Filtering and Preprocessing}
%\label{subsub:dataFilteringPreProcessing}

%The first step is the preprocessing task that includes among other activities, the data cleaning process where we perform outliers and noise removal \citep{andradeanomaly,gama2015extraccao}. 
%First of all, we need to look for duplicate data in the dataset and remove it.
%We also look for null data in the points where we cannot use the latitude or longitude to create new features in the next step.

%Due to the influence of GPS signal loss and data drift, there are a number of unwanted points in the trajectories set during the data acquisition. 
%Hence, cleaning tasks need to be performed in order to have more trustworthy data. 
%This inconsistent data must be deleted. We apply a smoothing median filter to each set of $5$ of GPS points to remove the noise as it is more robust to outliers \citep{andrade2020discovering,andrade2020mobility}.
%We also perform filtering according to the speed and acceleration: points with speed greater than $150 km/h$ and acceleration greater than $10 m/s2$ are removed~\citep{andrade2023estimating}.

%\subsubsection{Experimental Setup}

%For the clustering step, we set the $\epsilon$ to $100$ meters and the $MinPts$ to $5$. For the simplification of the trajectories using the RDP algorithm, we set the $\epsilon$ to $50$ due to the flexible moves the individuals may perform such as freely walking, running, cycling, or other activities related to vehicles.

\section{Conclusions}
In this work, we present a comprehensive large-scale study and implementation of analytics across various components of our project. We demonstrate the analysis of real-world spatiotemporal data to address practical problems, improve services, and drive advancements in urban environments. Our architecture is designed to meet industry needs and can handle vast amounts of data swiftly and effectively. We detail the extraction of trajectories, stay points, and preprocessing techniques essential for accurate data analysis.

Our research employs a range of machine learning models and algorithms, including rule-based models and transfer learning, showcasing the adaptability and reusability of these models across multiple use cases. We extract features such as average night and day stay times and length of stay, which can be incrementally updated to reflect changes in a user's residency or professional status. For the mode of transport identification, we propose a semi-supervised learning model based on pseudo-labeling and co-training, addressing the challenge of limited labeled data.

We identify anomalies in a city's normal spatial and temporal traffic flow by developing a metric that highlights deviations from typical traffic patterns. Additionally, we explore various techniques and distance measures to calculate frequent trajectories and discuss the suitability of different approaches in this context.

The practical applications derived from our models are encapsulated in several use cases, which include detailed statistics on geographical areas, such as user counts, visit frequency, time spent, and traffic flow between origins and destinations. The analysis also enabled the identification of different user profiles based on residency, profession, modes of transport, influence in catchment areas, and market share. These insights are visualized through interactive dashboards, providing stakeholders with valuable decision-making and strategic planning information for the selected geographical areas.

\section*{Acknowledgements}
The authors acknowledge the project CityAnalyser funded by the European Regional Development Fund (ERDF) through the Innovation and Digital Transition Programme (COMPETE 2020) under Portugal 2020 and by National Funds through the FCT - Fundação para a Ciência e a Tecnologia, I.P. (Portuguese Foundation for Science and Technology) with reference POCI-01-0247-FEDER-039924.

%\clearpage

\bibliography{paper,paper1}

@article{zheng2010geolife,
  title={Geolife: A collaborative social networking service among user, location and trajectory.},
  author={Zheng, Yu and Xie, Xing and Ma, Wei-Ying},
  journal={IEEE Data Eng. Bull.},
  volume={33},
  number={2},
  pages={32--39},
  year={2010},
  publisher={Citeseer}
}

@article{zheng2015trajectory,
  title={Trajectory data mining: an overview},
  author={Zheng, Yu},
  journal={ACM Transactions on Intelligent Systems and Technology (TIST)},
  volume={6},
  number={3},
  pages={1--41},
  year={2015},
  publisher={ACM New York, NY, USA}
}

@article{yuan2017review,
  title={A review of moving object trajectory clustering algorithms},
  author={Yuan, Guan and Sun, Penghui and Zhao, Jie and Li, Daxing and Wang, Canwei},
  journal={Artificial Intelligence Review},
  volume={47},
  pages={123--144},
  year={2017},
  publisher={Springer}
}

@inproceedings{ester1996density,
  title={A density-based algorithm for discovering clusters in large spatial databases with noise.},
  author={Ester, Martin and Kriegel, Hans-Peter and Sander, J{\"o}rg and Xu, Xiaowei and others},
  booktitle={Kdd},
  volume={96},
  number={34},
  pages={226--231},
  year={1996}
}

@article{andrade2020discovering,
  title={Discovering locations and habits from human mobility data},
  author={Andrade, Thiago and Cancela, Brais and Gama, Joao},
  journal={Annals of Telecommunications},
  volume={75},
  number={9},
  pages={505--521},
  year={2020},
  publisher={Springer}
}

@article{ramer1972iterative,
  title={An iterative procedure for the polygonal approximation of plane curves},
  author={Ramer, Urs},
  journal={Computer graphics and image processing},
  volume={1},
  number={3},
  pages={244--256},
  year={1972},
  publisher={Elsevier}
}

@article{douglas1973algorithms,
  title={Algorithms for the reduction of the number of points required to represent a digitized line or its caricature},
  author={Douglas, David H and Peucker, Thomas K},
  journal={Cartographica: the international journal for geographic information and geovisualization},
  volume={10},
  number={2},
  pages={112--122},
  year={1973},
  publisher={University of Toronto Press}
}

@inproceedings{chen2022semi,
  title={Semi-supervised learning with multi-head co-training},
  author={Chen, Mingcai and Du, Yuntao and Zhang, Yi and Qian, Shuwei and Wang, Chongjun},
  booktitle={Proceedings of the AAAI conference on artificial intelligence},
  volume={36},
  number={6},
  pages={6278--6286},
  year={2022}
}

@inproceedings{cascante2021curriculum,
  title={Curriculum labeling: Revisiting pseudo-labeling for semi-supervised learning},
  author={Cascante-Bonilla, Paola and Tan, Fuwen and Qi, Yanjun and Ordonez, Vicente},
  booktitle={Proceedings of the AAAI conference on artificial intelligence},
  volume={35},
  number={8},
  pages={6912--6920},
  year={2021}
}

@article{feng2016survey,
  title={A survey on trajectory data mining: Techniques and applications},
  author={Feng, Zhenni and Zhu, Yanmin},
  journal={IEEE Access},
  volume={4},
  pages={2056--2067},
  year={2016},
  publisher={IEEE}
}

@inproceedings{andrade2018identifying,
  title={Identifying points of interest and similar individuals from raw GPS data},
  author={Andrade, Thiago and Gama, Jo{\~a}o},
  booktitle={EAI International Conference on Smart Cities within SmartCity360° Summit},
  pages={293--305},
  year={2018},
  organization={Springer}
}

@article{andrade2020mobility,
  title={From mobility data to habits and common pathways},
  author={Andrade, Thiago and Cancela, Brais and Gama, Joao},
  journal={Expert Systems},
  volume={37},
  number={6},
  pages={e12627},
  year={2020},
  publisher={Wiley Online Library}
}

@inproceedings{andrade2020mining,
  title={Mining human mobility data to discover locations and habits},
  author={Andrade, Thiago and Cancela, Brais and Gama, Jo{\~a}o},
  booktitle={Machine Learning and Knowledge Discovery in Databases: International Workshops of ECML PKDD 2019, Proceedings Part II},
  pages={390--401},
  year={2020},
  organization={Springer}
}

@inproceedings{andrade2022you,
  title={How are you Riding? Transportation Mode Identification from Raw GPS Data},
  author={Andrade, Thiago and Gama, Jo{\~a}o},
  booktitle={EPIA Conference on Artificial Intelligence},
  pages={648--659},
  year={2022},
  organization={Springer}
}

@inproceedings{andrade2019discovering,
  title={Discovering Common Pathways Across Users’ Habits in Mobility Data},
  author={Andrade, Thiago and Cancela, Brais and Gama, Jo{\~a}o},
  booktitle={EPIA Conference on Artificial Intelligence},
  pages={410--421},
  year={2019},
  organization={Springer}
}

@inproceedings{andrade2024we,
  title={Where Do We Go From Here? Location Prediction from Time-Evolving Markov Models},
  author={Andrade, Thiago and Gama, Joao},
  booktitle={Proceedings of the 39th ACM/SIGAPP Symposium on Applied Computing},
  pages={365--367},
  year={2024}
}

@inproceedings{ye2009mining,
  title={Mining individual life pattern based on location history},
  author={Ye, Yang and Zheng, Yu and Chen, Yukun and Feng, Jianhua and Xie, Xing},
  booktitle={Mobile Data Management: Systems, Services and Middleware, 2009. MDM'09. Tenth International Conference on},
  pages={1--10},
  year={2009},
  organization={IEEE}
}

@inproceedings{vlachos2002DSM,
 author = {Vlachos, Michail and Gunopoulos, Dimitrios and Kollios, George},
 title = {Discovering Similar Multidimensional Trajectories},
 booktitle = {Proceedings of the 18th International Conference on Data Engineering},
 series = {ICDE '02},
 year = {2002},
 pages = {673--},
 url = {http://dl.acm.org/citation.cfm?id=876875.878994},
 urldate = {2019-06-16},
 acmid = {878994},
 publisher = {IEEE Computer Society},
 address = {Washington, DC, USA},
}

@article{eiter1994computing,
  title={Computing discrete Fr{\'e}chet distance},
  author={Eiter, Thomas and Mannila, Heikki},
  year={1994},
  journal={Technical Report CD-TR 94/64},
  publisher={Technische Universit\"{a}t Wien}
}

@inproceedings{magdy2015review,
  title={Review on trajectory similarity measures},
  author={Magdy, Nehal and Sakr, Mahmoud A and Mostafa, Tamer and El-Bahnasy, Khaled},
  booktitle={2015 IEEE seventh international conference on Intelligent Computing and Information Systems (ICICIS)},
  pages={613--619},
  year={2015},
  organization={IEEE}
}

@article{scikit-learn,
 title={Scikit-learn: Machine Learning in {P}ython},
 author={Pedregosa, F. and Varoquaux, G. and Gramfort, A. and Michel, V.
         and Thirion, B. and Grisel, O. and Blondel, M. and Prettenhofer, P.
         and Weiss, R. and Dubourg, V. and Vanderplas, J. and Passos, A. and
         Cournapeau, D. and Brucher, M. and Perrot, M. and Duchesnay, E.},
 journal={Journal of Machine Learning Research},
 volume={12},
 pages={2825--2830},
 year={2011}
}

@article{atluri2018spatio,
  title={Spatio-temporal data mining: A survey of problems and methods},
  author={Atluri, Gowtham and Karpatne, Anuj and Kumar, Vipin},
  journal={ACM Computing Surveys (CSUR)},
  volume={51},
  number={4},
  pages={1--41},
  year={2018},
  publisher={ACM New York, NY, USA}
}

@article{wang2020deep,
  title={Deep learning for spatio-temporal data mining: A survey},
  author={Wang, Senzhang and Cao, Jiannong and Philip, S Yu},
  journal={IEEE transactions on knowledge and data engineering},
  volume={34},
  number={8},
  pages={3681--3700},
  year={2020},
  publisher={IEEE}
}

@article{kraemer2020effect,
  title={The effect of human mobility and control measures on the COVID-19 epidemic in China},
  author={Kraemer, Moritz UG and Yang, Chia-Hung and Gutierrez, Bernardo and Wu, Chieh-Hsi and Klein, Brennan and Pigott, David M and Open COVID-19 Data Working Group† and Du Plessis, Louis and Faria, Nuno R and Li, Ruoran and others},
  journal={Science},
  volume={368},
  number={6490},
  pages={493--497},
  year={2020},
  publisher={American Association for the Advancement of Science}
}

@article{han2019cities,
  title={How do cities flow in an emergency? Tracing human mobility patterns during a natural disaster with big data and geospatial data science},
  author={Han, Su Yeon and Tsou, Ming-Hsiang and Knaap, Elijah and Rey, Sergio and Cao, Guofeng},
  journal={Urban Science},
  volume={3},
  number={2},
  pages={51},
  year={2019},
  publisher={MDPI}
}

@article{gerber2014predicting,
  title={Predicting crime using Twitter and kernel density estimation},
  author={Gerber, Matthew S},
  journal={Decision Support Systems},
  volume={61},
  pages={115--125},
  year={2014},
  publisher={Elsevier}
}

@article{cagnacci2010animal,
  title={Animal ecology meets GPS-based radiotelemetry: a perfect storm of opportunities and challenges},
  author={Cagnacci, Francesca and Boitani, Luigi and Powell, Roger A and Boyce, Mark S},
  journal={Philosophical Transactions of the Royal Society B: Biological Sciences},
  volume={365},
  number={1550},
  pages={2157--2162},
  year={2010},
  publisher={The Royal Society}
}

@article{steenbruggen2015data,
  title={Data from mobile phone operators: A tool for smarter cities?},
  author={Steenbruggen, John and Tranos, Emmanouil and Nijkamp, Peter},
  journal={Telecommunications Policy},
  volume={39},
  number={3-4},
  pages={335--346},
  year={2015},
  publisher={Elsevier}
}

@article{mahrez2021smart,
  title={Smart urban mobility: When mobility systems meet smart data},
  author={Mahrez, Zineb and Sabir, Essaid and Badidi, Elarbi and Saad, Walid and Sadik, Mohamed},
  journal={IEEE Transactions on Intelligent Transportation Systems},
  volume={23},
  number={7},
  pages={6222--6239},
  year={2021},
  publisher={IEEE}
}

@article{zhu2018big,
  title={Big data analytics in intelligent transportation systems: A survey},
  author={Zhu, Li and Yu, Fei Richard and Wang, Yige and Ning, Bin and Tang, Tao},
  journal={IEEE Transactions on Intelligent Transportation Systems},
  volume={20},
  number={1},
  pages={383--398},
  year={2018},
  publisher={IEEE}
}

@article{liao2005clustering,
  title={Clustering of time series data—a survey},
  author={Liao, T Warren},
  journal={Pattern recognition},
  volume={38},
  number={11},
  pages={1857--1874},
  year={2005},
  publisher={Elsevier}
}

@inproceedings{yue2019detect,
  title={Detect: Deep trajectory clustering for mobility-behavior analysis},
  author={Yue, Mingxuan and Li, Yaguang and Yang, Haoze and Ahuja, Ritesh and Chiang, Yao-Yi and Shahabi, Cyrus},
  booktitle={2019 IEEE International Conference on Big Data (Big Data)},
  pages={988--997},
  year={2019},
  organization={IEEE}
}

@article{lv2016big,
  title={Big data driven hidden Markov model based individual mobility prediction at points of interest},
  author={Lv, Qiujian and Qiao, Yuanyuan and Ansari, Nirwan and Liu, Jun and Yang, Jie},
  journal={IEEE Transactions on Vehicular Technology},
  volume={66},
  number={6},
  pages={5204--5216},
  year={2016},
  publisher={IEEE}
}

@inproceedings{sun2017dxnat,
  title={DxNAT—Deep neural networks for explaining non-recurring traffic congestion},
  author={Sun, Fangzhou and Dubey, Abhishek and White, Jules},
  booktitle={2017 IEEE International Conference on Big Data (Big Data)},
  pages={2141--2150},
  year={2017},
  organization={IEEE}
}

@article{ouyang2016deepspace,
  title={Deepspace: An online deep learning framework for mobile big data to understand human mobility patterns},
  author={Ouyang, Xi and Zhang, Chaoyun and Zhou, Pan and Jiang, Hao and Gong, Shimin},
  journal={arXiv preprint arXiv:1610.07009},
  year={2016}
}

@article{wang2020predictability,
  title={Predictability and prediction of human mobility based on application-collected location data},
  author={Wang, Huandong and Zeng, Sihan and Li, Yong and Jin, Depeng},
  journal={IEEE Transactions on Mobile Computing},
  volume={20},
  number={7},
  pages={2457--2472},
  year={2020},
  publisher={IEEE}
}

@inproceedings{kong2018hst,
  title={HST-LSTM: A hierarchical spatial-temporal long-short term memory network for location prediction.},
  author={Kong, Dejiang and Wu, Fei},
  booktitle={IJCAI},
  volume={18},
  pages={2341--2347},
  year={2018}
}

@article{liu2021attention,
  title={An attention-based category-aware GRU model for the next POI recommendation},
  author={Liu, Yuwen and Pei, Aixiang and Wang, Fan and Yang, Yihong and Zhang, Xuyun and Wang, Hao and Dai, Hongning and Qi, Lianyong and Ma, Rui},
  journal={International Journal of Intelligent Systems},
  volume={36},
  number={7},
  pages={3174--3189},
  year={2021},
  publisher={Wiley Online Library}
}

@article{xu2021towards,
  title={Towards a multidimensional view of tourist mobility patterns in cities: A mobile phone data perspective},
  author={Xu, Yang and Xue, Jiaying and Park, Sangwon and Yue, Yang},
  journal={Computers, Environment and urban systems},
  volume={86},
  pages={101593},
  year={2021},
  publisher={Elsevier}
}

@article{sole2016model,
  title={A model to identify urban traffic congestion hotspots in complex networks},
  author={Sol{\'e}-Ribalta, Albert and G{\'o}mez, Sergio and Arenas, Alex},
  journal={Royal Society open science},
  volume={3},
  number={10},
  pages={160098},
  year={2016},
  publisher={The Royal Society}
}

@article{islam2022survey,
  title={A survey on deep learning based Point-of-Interest (POI) recommendations},
  author={Islam, Md Ashraful and Mohammad, Mir Mahathir and Das, Sarkar Snigdha Sarathi and Ali, Mohammed Eunus},
  journal={Neurocomputing},
  volume={472},
  pages={306--325},
  year={2022},
  publisher={Elsevier}
}

@article{huang2019transport,
  title={Transport mode detection based on mobile phone network data: A systematic review},
  author={Huang, Haosheng and Cheng, Yi and Weibel, Robert},
  journal={Transportation Research Part C: Emerging Technologies},
  volume={101},
  pages={297--312},
  year={2019},
  publisher={Elsevier}
}

@article{dabiri2018inferring,
  title={Inferring transportation modes from GPS trajectories using a convolutional neural network},
  author={Dabiri, Sina and Heaslip, Kevin},
  journal={Transportation research part C: emerging technologies},
  volume={86},
  pages={360--371},
  year={2018},
  publisher={Elsevier}
}

@article{ben2019clustering,
  title={Clustering users by their mobility behavioral patterns},
  author={Ben-Gal, Irad and Weinstock, Shahar and Singer, Gonen and Bambos, Nicholas},
  journal={ACM Transactions on Knowledge Discovery from Data (TKDD)},
  volume={13},
  number={4},
  pages={1--28},
  year={2019},
  publisher={ACM New York, NY, USA}
}

@article{fanaee2016event,
  title={Event detection from traffic tensors: A hybrid model},
  author={Fanaee-T, Hadi and Gama, Joao},
  journal={Neurocomputing},
  volume={203},
  pages={22--33},
  year={2016},
  publisher={Elsevier}
}

@article{dolega2016estimating,
  title={Estimating attractiveness, hierarchy and catchment area extents for a national set of retail centre agglomerations},
  author={Dolega, Les and Pavlis, Michalis and Singleton, Alex},
  journal={Journal of Retailing and Consumer Services},
  volume={28},
  pages={78--90},
  year={2016},
  publisher={Elsevier}
}

@article{daniotti2023maximum,
  title={A maximum entropy approach for the modelling of car-sharing parking dynamics},
  author={Daniotti, Simone and Monechi, Bernardo and Ubaldi, Enrico},
  journal={Scientific Reports},
  volume={13},
  number={1},
  pages={2993},
  year={2023},
  publisher={Nature Publishing Group UK London}
}

@article{cronje2020review,
  title={A review on tourism destination competitiveness},
  author={Cronj{\'e}, Dani{\'e}lle Francoise and du Plessis, Engelina},
  journal={Journal of Hospitality and Tourism Management},
  volume={45},
  pages={256--265},
  year={2020},
  publisher={Elsevier}
}

@article{drezner2023multipurpose,
  title={Multipurpose shopping trips and location},
  author={Drezner, Tammy and O’Kelly, Morton and Drezner, Zvi},
  journal={Annals of Operations Research},
  volume={321},
  number={1-2},
  pages={191--208},
  year={2023},
  publisher={Springer}
}

@article{pratt2014delineating,
  title={Delineating retail conurbations: A rules-based algorithmic approach},
  author={Pratt, Matthew D and Wright, Jim A and Cockings, Samantha and Sterland, Iain},
  journal={Journal of Retailing and Consumer Services},
  volume={21},
  number={5},
  pages={667--675},
  year={2014},
  publisher={Elsevier}
}

@inproceedings{kohani2012exact,
  title={Exact approach to the tariff zones design problem in public transport},
  author={Koh{\'a}ni, Michal},
  booktitle={Proceedings of the International Conference Mathematical Methods in Economics},
  volume={30},
  year={2012}
}

@article{ma2022individual,
  title={Individual mobility prediction review: Data, problem, method and application},
  author={Ma, Zhenliang and Zhang, Pengfei},
  journal={Multimodal transportation},
  volume={1},
  number={1},
  pages={100002},
  year={2022},
  publisher={Elsevier}
}

@inproceedings{aljeri2020performance,
  title={A performance evaluation of time-series mobility prediction for connected vehicular networks},
  author={Aljeri, Noura and Boukerche, Azzedine},
  booktitle={Proceedings of the 16th ACM Symposium on QoS and Security for Wireless and Mobile Networks},
  pages={127--131},
  year={2020}
}

@article{hou2016predicting,
  title={Predicting mobile users’ behaviors and locations using dynamic Bayesian networks},
  author={Hou, Jianrong and Zhao, Hui and Zhao, Xiaofeng and Zhang, Jie},
  journal={Journal of Management Analytics},
  volume={3},
  number={3},
  pages={191--205},
  year={2016},
  publisher={Taylor \& Francis}
}

@article{bahadori2014fast,
  title={Fast multivariate spatio-temporal analysis via low rank tensor learning},
  author={Bahadori, Mohammad Taha and Yu, Qi Rose and Liu, Yan},
  journal={Advances in neural information processing systems},
  volume={27},
  year={2014}
}

@inproceedings{fattore2020automec,
  title={AutoMEC: LSTM-based user mobility prediction for service management in distributed MEC resources},
  author={Fattore, Umberto and Liebsch, Marco and Brik, Bouziane and Ksentini, Adlen},
  booktitle={Proceedings of the 23rd International ACM Conference on Modeling, Analysis and Simulation of Wireless and Mobile Systems},
  pages={155--159},
  year={2020}
}

@inproceedings{djukic2012efficient,
  title={Efficient real time OD matrix estimation based on Principal Component Analysis},
  author={Djukic, Tamara and Fl{\"o}tter{\"o}d, Gunnar and Van Lint, Hans and Hoogendoorn, Serge},
  booktitle={2012 15th International IEEE Conference on Intelligent Transportation Systems},
  pages={115--121},
  year={2012},
  organization={IEEE}
}

@article{krishnakumari2019data,
  title={A data driven method for OD matrix estimation},
  author={Krishnakumari, Panchamy and van Lint, Hans and Djukic, Tamara and Cats, Oded},
  journal={Transportation Research Procedia},
  volume={38},
  pages={139--159},
  year={2019},
  publisher={Elsevier}
}

@article{moreira2016time,
  title={Time-evolving OD matrix estimation using high-speed GPS data streams},
  author={Moreira-Matias, Lu{\'\i}s and Gama, Jo{\~a}o and Ferreira, Michel and Mendes-Moreira, Jo{\~a}o and Damas, Luis},
  journal={Expert systems with Applications},
  volume={44},
  pages={275--288},
  year={2016},
  publisher={Elsevier}
}

@article{ou2019learn,
  title={Learn, assign, and search: real-time estimation of dynamic origin-destination flows using machine learning algorithms},
  author={Ou, Jishun and Lu, Jiawei and Xia, Jingxin and An, Chengchuan and Lu, Zhenbo},
  journal={IEEE Access},
  volume={7},
  pages={26967--26983},
  year={2019},
  publisher={IEEE}
}

@inproceedings{shaji2022study,
  title={Study on Correlation Between Vehicle Emissions and Air Quality in Porto},
  author={Shaji, Nirbhaya and Andrade, Thiago and Ribeiro, Rita P and Gama, Jo{\~a}o},
  booktitle={Joint European Conference on Machine Learning and Knowledge Discovery in Databases},
  pages={181--196},
  year={2022},
  organization={Springer}
}

@misc{gcp,
title={Google Cloud Services},
url={https://cloud.google.com/?hl=en},
author={Google},
publisher={Google},
year={Accessed on Mar 18, 2024}
}

@misc{dataflow,
title={Google Cloud Services},
url={https://cloud.google.com/dataflow/},
author={Google},
publisher={Google},
year={Accessed on Mar 18, 2024}
}

@misc{bigquery,
title={Google Cloud Services},
url={https://cloud.google.com/bigquery/},
author={Google},
publisher={Google},
year={Accessed on Mar 18, 2024}
}

@misc{vertexAI,
title={Google Cloud Services},
url={https://cloud.google.com/vertex-ai/},
author={Google},
publisher={Google},
year={Accessed on Mar 18, 2024}
}

@misc{powerbi,
title={Microsoft Power BI},
url={https://www.microsoft.com/en-us/power-platform/products/power-bi/},
author={Microsoft},
publisher={Microsoft},
year={Accessed on Mar 18, 2024}
}

\end{document}